\documentclass{article}
\usepackage[english]{babel}
\usepackage{ctable}
\usepackage{amsmath,amssymb,amsthm}
\usepackage{enumerate}
\usepackage{enumitem}

\usepackage[linesnumbered,ruled,vlined]{algorithm2e}
\usepackage[noend]{algorithmic}
\usepackage{hyperref}
\usepackage{authblk} %author affiliations
\usepackage[nottoc]{tocbibind} % per aggiungere la bibliografia come voce ToC
\newtheorem{theorem}{Theorem}

\newtheorem{definition}{Definition}%[section]
\newtheorem{proposition}{Proposition}

\DeclareMathOperator*{\argmin}{argmin}
\DeclareMathOperator*{\argmax}{argmax}
\DeclareMathOperator*{\supp}{supp}
\DeclareMathOperator{\diag}{diag}

\newcommand{\bb}{\mathbb}
\newcommand{\R}{\mathbb{R}}
\newcommand{\N}{\mathbb{N}}

\newcommand{\ESNR}{\text{E}_\text{SNR}}
\renewcommand{\a}{\alpha}
\graphicspath{{images/}}

\title{Sparse Models for Machine Learning}
\author{Jianyi Lin}
\affil{Department of Statistical Sciences\\Università Cattolica del Sacro Cuore\\Milan, Italy\\jianyi.lin@unicatt.it}

\date{}

\begin{document}
\maketitle
\tableofcontents
\newpage
\begin{abstract}
%\addcontentsline{toc}{section}{Abstract}
Arguably one of the most notable forms of the principle of parsimony was formulated by the philosopher and theologian William of Ockham in the 14th century, and later became well known as Ockham's Razor principle, which can be phrased as: ``Entities should not be multiplied without necessity.'' This principle is undoubtedly one of the most fundamental ideas that pervade many branches of knowledge, from philosophy to art and science, from ancient times to modern age, then summarized in the expression ``Make everything as simple as possible, but not simpler'' as likewise asserted Albert Einstein.

The sparse modeling is an evident manifestation capturing the parsimony principle just described, and sparse models are widespread in statistics, physics, information sciences, neuroscience, computational mathematics, and so on.
In statistics the many applications of sparse modeling span regression, classification tasks, graphical model selection, sparse M-estimators and sparse dimensionality reduction. It is also particularly effective in many statistical and machine learning areas where the primary goal is to discover predictive patterns from data which would enhance our understanding and control of underlying physical, biological, and other natural processes, beyond just building accurate outcome black-box predictors. Common examples include selecting biomarkers in biological procedures, finding relevant brain activity locations which are predictive about brain states and processes based on fMRI data, and identifying network bottlenecks best explaining end-to-end performance.

Moreover, the research and applications of efficient recovery of high--di\-men\-sio\-nal sparse signals from a relatively small number of observations, which is the main focus of compressed sensing or compressive sensing \cite{eldar2012compressed,donoho2006compressed}, have rapidly grown and became an extremely intense area of study beyond classical signal processing.
Likewise interestingly, sparse modeling is directly related to various artificial vision tasks, such as image denoising \cite{eladaharon06}, segmentation, restoration and superresolution \cite{mairal2014sparse,cheng2015sparse}, object or face detection and recognition in visual scenes \cite{adamo2015robust,grossi2016robust}, as well as action recognition and behavior analysis \cite{guha2011learning}. Sparsity has also been applied in information compression \cite{grossi2015high}, text classification and recommendation systems \cite{ning2011slim}.

In this manuscript, we provide a brief introduction of the basic theory underlying sparse representation and compressive sensing, and then discuss some methods for recovering sparse solutions to optimization problems in effective way, together with some applications of sparse recovery in a machine learning problem known as sparse dictionary learning.
\end{abstract}

\section{Introduction}
We start with presenting the sparsity from a signal perspective following the approach in \cite{eldar2012compressed}.
Shannon-Nyquist sampling theorem \index{Shannon-Nyquist sampling theorem} is one of the central principle in classical signal processing. For a lossless reconstruction of a continuous-time signal $s(t)$ having harmonics with no frequencies higher than $B>0$ Hertz from the signal samples, it is sufficient to sample $s(t)$ at a regular rate $A > 2B$.
But in the last couple of decades the studies in an emerging field now known as \textit{compressed sensing} \index{compressed sensing} or compressive sensing (CS) have advanced beyond the Shannon-Nyquist limits for signal acquisition and sensor design \cite{Duarte08,Chartrand2007}, showing that a signal can be reconstructed from far fewer measurements than what is classically considered necessary, provided that it admits a compressible or sparse representation.
Instead of taking $n$ signal samples at a regular period, in CS one performs the measurements through dot products with $p \ll n$ measurement vectors of $\R^n$, that represent the characteristics of the phenomenon sensing process, and then recovers the signal via sparsity promoting optimization methods. In matrix notation, the measures $y$ can be expressed as $y = \Psi s$ where the rows of the $p \times n$ matrix $\Psi$ contain the measurement vectors and $s$ is the sampled signal.

In this setting, it is common to consider $s$ as sparse, or alternatively sparsely representable, when
$$
s=\Phi\alpha
$$
for some orthogonal matrix $\Phi\in\R^{n\times n}$, where $\alpha$ is a sparse encoding, e.g.\ a truncated transform-based coding. While the matrix $\Psi \Phi$ might be rank-deficient, and hence its corresponding measurement procedure loses information in general, it can be shown however that it preserves the information in sparse and compressible signals under a notable range of conditions; one typical example is represented by the Restricted Isometry Property (RIP) \cite{Tanner2011} of order $2k$, from which the standard CS theory ensures very likely a robust signal recovery from $p = \mathcal{O}(k \log \frac{n}{k})$ measurements.
Moreover, many fundamental works developed by Cand\'{e}s, Chen, Saunders, Tao and Romberg 
\cite{CDS98,Tao06,Candes06,Candes06b,Candes06c} converge to the evidence that a finite dimensional 
signal having a sparse or compressible representation can be recovered exactly from a small set of linear non adaptive
measurements.

This chapter starts with some preliminary notions in linear algebra and proceed with an introduction to the sparse optimization problem and recall some of the most important 
results in literature that summarize conditions under which the sparse recovery algorithms later introduced are able to recover 
the sparsest representation of a signal under a given frame or dictionary. The design, through machine learning, of well representative frames will be the subject of interest in the ending part of the chapter dedicated to applications.

\section {Sparse Vectors}
The key point in the brief introduction above is of course what it is deemed as sparse, since this is undoubtedly the most clear and prominent form of parsimony.
A first significant definition of sparsity for a vector we introduce simply counts the number of non-null entries.

Consider a vector $x \in \R^n$ and define the functional $\|x\|_p=(\sum_{i=1}^{m} |x_i|^p)^{1/p}$; it is known that this functional is a norm for $p\geq 1$, called $\ell_p$-norm or $p$-norm\footnote{The $1$-norm and $2$-norm are the well known Manhattan norm and Euclidean norm, respectively.}, and so it is in the limit case $\|x\|_\infty= \lim_{p\rightarrow\infty} \|x\|_p=\max\{|x_i|:i=1,...,n\}$, called uniform norm or max norm. %, while it is a quasi-norm \cite{benyaminilindenstrauss91} for $0<p<1$. 
If $0 < p < 1$, $\| . \|_p$ is a quasinorm \index{quasinorm} \cite{benyaminilindenstrauss91}, i.e. it satisfies the axioms of the norm except the triangle inequality, which is replaced by the quasi-triangle inequality
\begin{equation}
\label{QUASINORM}
\| x + y \|_p \leq \gamma \left ( \| x  \|_p + \| y \|_p \right ) 
\end{equation}
for some $\gamma \geq 1$, the smallest of which is called the quasinorm's constant. A vector space with an associated quasinorm is called a quasinormed vector space\index{quasinormed vector space}.

\begin{figure}[h]
	\centering
	\includegraphics[width=\textwidth]{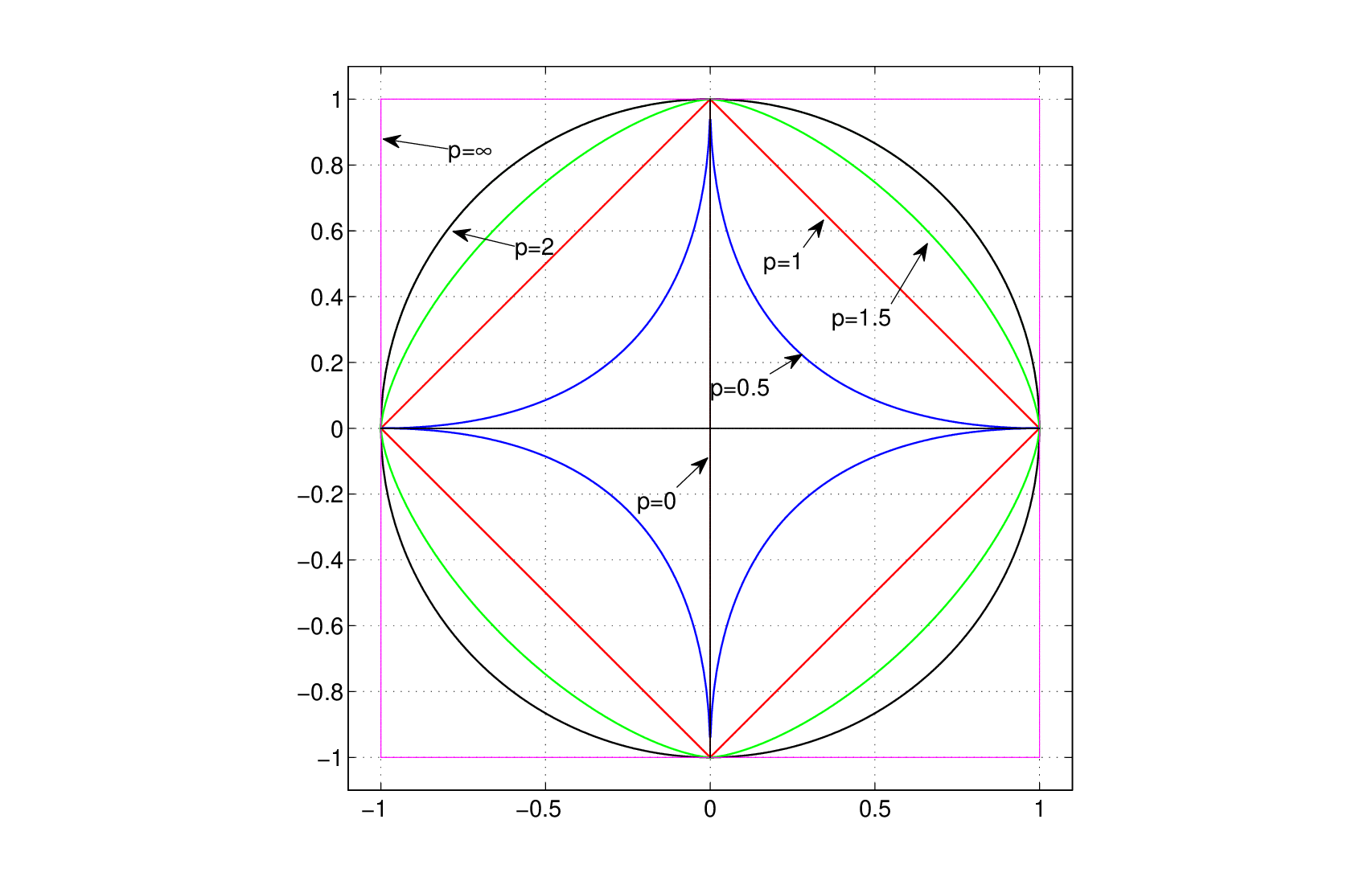}
	\caption[Unit spheres]{Unit balls in $\R^2$ endowed with the $p$-norms with $p = 1, 2, \infty$, the $p$-quasinorm with $p = 0.5$ and the $0$-pseudonorm.}
	\label{fig:lpnorm}
\end{figure}

The support of $x$ is defined by $\supp(x)=\{i:x_i\neq 0\}$. The functional
$$
\|x\|_0 := \sum_{i=1}^n \mathbf{1}(x_i\neq 0) = \lim_{q\downarrow 0} \|x\|_q^q
$$
satisfies the triangle inequality but not the absolute homogeneity condition, stated as $\forall \lambda\in\R,x\in\R^n: \|\lambda x\| =|\lambda|\|x\|$, and hence is called a pseudo-norm; nevertheless it is often referred to improperly as $0$-norm or $0$-quasinorm as well, and we will keep this slight abuse of language. This pseudo-norm is the main measure of sparsity.
In Figure \ref{fig:lpnorm} some unit balls $\{ x  : \| x \|_p \leq 1 \}$ are depicted on the plane endowed with $\| \cdot \|_0$, some norms and quasinorms for different values of $p$. We see that the convexity holds only for $p\geq 1$.

The vector $x$ is $k$-\textit{sparse} \index{sparse vector} when it has at most $k$ non-null entries, i.e.$\| x \|_0 \leq k$, and we denote the set of all $k$-sparse vectors with $\Sigma_k = \{ x : \| x \|_0 \leq k \}$.
In the real world, rarely the signals are truly sparse, rather they can be considered {compressible} \index{compressible signal} in the sense of good approximation by a sparse signal.
We can quantify the compressibility of a signal $s$ through the $\ell_p$ error $\sigma_k(s)_p$ between the original signal and best $k$-term approximation in $\Sigma_k$:
\begin{equation}
 \sigma_k(s)_p = \inf_{\hat{s} \in \Sigma_k} \| s - \hat{s}\|_p \qquad \text{for }p>0.
\end{equation}
For $k$-sparse vectors $s \in \Sigma_{k}$ of course $\sigma_k(s)_p = 0$ for any $p$.

Moreover, a compressible or sparse signal $s=\Phi\a$ corresponds also to a fast rate decay of the coefficient magnitude sequence $\{|\alpha_i|\}$ sorted in descending order, so that they can be represented accurately by $k \ll m$ coefficients \cite{eldar2012compressed}. For such kind of signals there exist constants $C$, $r > 0$ such that 
$$
	\sigma_k(s)_2 \leq C k^{-r}.
$$
In that case, one can show that the sparse approximation error $\sigma_k(s)_2$ will shrink as $k^{-r}$ if and only if the sorted coefficients $\{|\alpha_i|\}$ have a decay rate $i^{-r+\frac{1}{2}}$ \cite{DeVore98}.

\section {Sparse Solutions to Underdetermined Systems}
The pursue of a sparse source signal from some measurement hence corresponds to finding a $k$-sparse solution $\a$ of a linear system of the kind $s=\Phi \a$, for some integer $k>0$.
Finding sparse solutions of underdetermined systems of linear equations is a topic extensively studied \cite{D04,CDS98,Candes2005}, and many problems across different disciplines rely on advantages from finding sparse solutions. In general, all these tasks amount to solving the problem $\Phi\alpha=s$ with a $n\times m$ matrix $\Phi$ and $n<m$. Depending on the various application contexts, $\Phi=[\phi_1,\dots,\phi_m]$ is a collection of $m$ vectors in $\mathbb{R}^n$ representing basic waveforms, usually called atoms, and the matrix $\Phi$ is called frame or dictionary\footnote{The latter is used more often in computer science or engineering areas.}, which is formally defined as a collection of (column) vectors $\phi_i\in\R^n$ such that
$$
a\|x\|^2\leq \|\Phi x\|^2\leq b\|x\|^2 \quad \text{ for all } x\in\R^n
$$
for some $0<a\leq b <\infty$. These two constants are the so-called \textit{frame bounds}, which are in fact the least and the greatest singular value of $\Phi$: $a=\sigma_n(\Phi)$ and $b=\sigma_1(\Phi)$, respectively. The transpose Moore-Penrose pseudoinverse $(\Phi^\dagger)^T$ is the so-called canonical dual frame, which is still a frame for $\R^n$ with frame bounds $0<\frac{1}{b}\leq \frac{1}{a}< \infty$ \cite[Theor. 5.5]{Mallat08}. From the definition it is clear that a frame has full rank since the smallest singular value must be positive, and moreover, having assumed that $n<m$, a frame is said to be ``overcomplete'' since it contains more elements than a basis. As definition, a frame is said to be tight when $a=b$ and this occurs exactly when the non-null eigenvalues of the Gram matrix $G=\Phi^T\Phi$ are all the same. We have a Parseval frame when $a=b=1$. An equiangular frame is a collection $\Phi=[\phi_1, ...,\phi_m]$ of equal-norm vectors spanning the space $\R^n$, such that any pairwise dot product has the same magnitude, i.e. $|\langle \phi_i,\phi_j\rangle| = \theta$ for $i\neq j$. The equiangular frames $\Phi$ that are unit-norm and tight are called equiangular tight frames (ETFs) or optimal Grassmannian frames, and in such cases the common angle between atoms is described by the condition $\theta = \sqrt{\frac{m-n}{n(m-1)}}$ \cite{Datta2021}. This special value is referred to as Welch bound since it appears in the inequality
$$
\mu(\Phi) := \max_{i\neq j} |\langle \phi_i,\phi_j\rangle| \geq \sqrt{\frac{m-n}{n(m-1)}}
$$
established by Welch in \cite{welch1974lower} for general unit-norm frames. The dual of an ETF is an ETF too. The existence of an ETF is not guaranteed for every pair $(n,m)$ \cite{sustik2007existence}, but the effective construction of ETFs or their approximations \cite{elad2010} are particularly of interest in data representation models since the dictionary attaining the Welch bound has atoms uniformly spanning the space that hence allow for easily encoding the data points.
More practically, the dictionary generally provides a redundant way of representing a signal in $\R^n$.

With the above premises, the overcomplete dictionary $\Phi$ leads to $\infty^{m-n}$ many solutions of the system $\Phi \a=s$ corresponding to the coefficients of as many linear combinations of the atoms in $\Phi$ for representing $s$. Such kind of systems lacking uniqueness in the solution typically represent inverse problems in science and engineering that are ill-posed in Hadamard sense.
In ill-posed problems, we desire to build a single solution of $s = \Phi \alpha$ by introducing some additional identifying criteria.
To this aim, a classical approach is the \textit{regularization} technique, for which one of the earliest representatives is Tikhonov's regularization \cite{tikhonov1943stability}. In regularization techniques, a function $J(\alpha)$ that evaluates the desirability of a would-be solution $\alpha$ is introduced, with smaller values being preferred.
Indeed, by formulating the general optimization \index{regularization problem} problem 
\begin{equation}
\label{PJ}
\tag{PJ}
\min_{\alpha \in \R^m}\ J( \alpha ) \quad \mbox{ subject to }\ \Phi \alpha = s 
\end{equation}
one wants to reconstruct one and possibly the only solution $\hat \alpha \in \R^m$ of the linear system that enjoys an optimal value w.r.t. the desirability quantified by $J$.

One of those desirable qualities can be given by the sparsity norm $J(\alpha) = \| \alpha \|_0$ of the solution. Therefore, the sparse recovery problem, where the goal is to recover a high-dimensional vector $\alpha$ with few non-null entries from an observation $s$, can be formalized into the optimization problem 
\begin{equation}\label{P0}
\min_{\alpha\in\mathbb{R}^m}\|\alpha\|_0\quad \text{subject to} \quad \Phi\alpha=s.\tag{$P_0$}
\end{equation}
Tackling the non-convex problem (\ref{P0}) naively entails the searches over almost all $2^m$ subsets of columns of $\Phi$ corresponding to non-null positions of $\a$, a procedure which is clearly combinatorial in nature and has high computational complexity. Indeed, (\ref{P0}) was proved to be NP-hard \cite{Natarajan95}.

Another early choice for a regularization approach is through the Euclidean norm $J(\a) = \|\a\|_2$. This special case admits the well-known unique solution $\alpha_{LS}$ that can be written in closed-form
\begin{equation}
\label{PSEUDOINVERSE_LS}
\alpha_{LS} =  \Phi^{\dagger} s = \Phi^T(\Phi \Phi^T)^{-1} s.
\end{equation}
Indeed, it is straight-forward to show that $\alpha_{LS}$ in (\ref{PSEUDOINVERSE_LS}) has $\ell_2$ norm bounding below all the vectors $\a$ satisfying $\Phi\a=s$:
\begin{equation}
\label{LEAST_SQUARE_SOL}
\| \alpha_{LS}\|^2_2 \leq \| \alpha \|^2_2
\end{equation}
and therefore is called the \textit{least squares} solution\index{least squares solution}.

The 0-norm and the Euclidean norm correspond somewhat to two extreme choices for the regularization based on the family of $\ell_p$ (pseudo/quasi)norms. The two cases actually spans a range of intermediate techniques introduced for inducing sparsity or controlling the regularization of the solution, so the following section is dedicated to outline some of those relevant methods from a statistical perspective. We will notice that, contrarily to the system of equalities introduced in this section, those models in statistical inference naturally admits some desirably low error between $\Phi\a$ and $s$, while keeping a trade-off with the goal of sparsity.

The sparse recovery problem \ref{P0} \cite{Candes08a,Candes08b} can also be relaxed to the convex $\ell_1$ based problem
\begin{equation}\label{P1}
	\min_{\alpha \in \R^m}\ \| \alpha \|_{1} \mbox{ s.t. } \Phi \alpha = s 	\tag{$P_1$}
\end{equation}
where $\| \alpha \|_{1} = \sum_{i=1}^m | \alpha_i |$ is the $\ell_1$ norm of vector $\alpha$. This can be reformulated as a linear program (LP) \cite{Rudelson06} 
\begin{equation}
\label{LP}
\min_{ t \in \R^m } \sum_{ i=1 }^m t_i \mbox{ s.t. } -t \leq \alpha \leq t, \Phi \alpha = s
\end{equation}
with inequalities on the vector variables $\a$ and $t$ to be understood element-wise.
This problem can be solved exactly with classical tools such as interior point methods or the simplex algorithm, although the linear programming \index{linear programming} formulation (\ref{LP}) has the drawback of computational inefficiency in most cases. For this reason other dedicated algorithms aimed at directly solving \ref{P1} have been proposed in literature: for example, the greedy Basis Pursuit \index{sparse recovery algorithm!greedy!Basis Pursuit} (BP) \cite{CDS98}, or the Least Angle Regression 
\index{sparse recovery algorithm!relaxation!Least Angle Regression} (LARS) \cite{Efron2004}.

The relationship between the above introduced problems will be illustrated on the basis of properties concerning the sensing matrix $\Phi$ in the next sections, after a short digression on the connections with sparse statistical models.

\section{Sparse Statistical Models}
The formulation of some inference procedure on statistical models, such as regression models, that adheres to some parsimony or low-complexity principle is typically rephrased as a problem of loss function minimization with some regularization-based constraint, as the following kind
\begin{equation}\label{prob:constrained_minloss}
\min_\beta\ L(\beta; Z, \bb D) \quad \text{subject to}\quad J(\beta)\leq t
\end{equation}
where $(\bb D,Z)$ represents the data from the predictor and response variable pair, and $\beta$ is the parameter vector of the model. In many of these procedures, such as maximum likelihood or ordinary least squares estimation with sparsity, the minimization problem above boils down to the $\ell_0$ constrained formulation
\begin{equation}\tag{SAP}\label{eq:SAP}
\underset{\beta\in\R^p}{\min}  \|Y - \bb X\beta\|_2^2 \quad \text{ subject to} \quad  \|\beta\|_0 \leq t
\end{equation}
with $\bb X \in \R^{n\times p}$, but of course other choices are suitable for sparse inference methods as we will see now.

One of the earliest methods studied is Lasso: least absolute shrinkage and selection operator.
The Lasso \cite{hastie2017}, also known as \textit{basis pursuit} in computer science community, solves a convex relaxation of \ref{eq:SAP} where the $\ell_0$-norm is replaced by the total absolute value of the parameters $\|\beta\|_1=\sum_i |\beta_i|$, namely
\begin{equation}\tag{Lasso}\label{prob:Lasso}
\underset{\beta\in\R^p}{\min}  \|Y - \bb X\beta\|_2^2 \quad \text{ subject to } \|\beta\|_1\leq t
\end{equation}
where $t>0$ is a parameter representing a ``budget'' for how well the data can be fitted, since a shrunken parameter estimate corresponds to a more heavily constrained model \cite{hastie2015}. This hyper-parameter is usually tuned by cross-validation. In general, a Lasso estimator $\hat\beta_L$ is a biased estimator of the true value vector $\beta$, and the bias $\mathbb E(\hat\beta_L - \beta)$ could be arbitrarily large depending on the value of the constraint threshold $t$.

As optimization problem, Lasso is a convex problem and may have non-unique solution whenever the predictor variables are collinear. It does not admit a closed-form solution, but nevertheless it can be efficiently solved by studying its equivalent Lagrangrian function form $\min_\beta \|Y - \bb X\beta\|_2^2 + \lambda\|\beta\|_1$, also known as basis pursuit \textit{denoising}\index{denoising} (BPDN), and then applying non-smooth unconstrained optimization techniques, e.g. coordinate descent methods or resorting to the proximal Newton map method, which has also been used for addressing the $\ell_1$ sparse logistic regression \cite{hastie2015}.
%Comparing to the latter method, we notice that our approach does not convert the sparse fitting problem to a convex relaxation, but rather maintains the non-convex $\ell_0$-norm based constraint, and hence our algorithm does not use the penalization techniques for the likelihood function in the formulation.

Notice that this Lagrangian formulation corresponds to adding a \textit{penalization}\index{penalization} term to the original objective function that hinders the large magnitude parameter vectors $\beta$, which is the approach of penalty methods \cite{bertsekas2016nonlinear} for turning constrained optimization into an unconstrained form.
Since in Lasso the $\ell_0$-norm is replaced with the $\ell_1$-norm, the estimate $\hat\beta_L$ differs from the SAP solutions in general, but nevertheless the recovery of truly sparse parameter vector $\beta$ is feasible when some classical conditions on the matrix $\bb X$ are satisfied, such as the ones we will introduce in the following sections: the Nullspace Property, which is guaranteed in turn by the Restricted Isometry Property or a sufficiently bounded Mutual Coherence \cite{elad2010}.

Among the other penalization approaches to address the sparse regression, the \textit{elastic net} method lies in between the Lasso formulation and the ridge regression \cite{zou2005regularization}, the latter being the statistical counterpart of traditional Tikhonov regularization techniques for coping with ill-conditioned data in differential problems, specifically introduced in mathematical physics in early years \cite{tikhonov1943stability}. Adopting the linear combination of Lasso $\ell_1$ term and $\ell_2$ ridge penalty term in the objective function, the elastic net deals better with predictor variables that are correlated and tends to group correlated features, hence promoting a basic form of structured sparsity \cite{Yuan2006Model}. Indeed, this can mitigate the erratic behavior of the $\hat\beta_i$ coefficient estimate as result of adding the ridge penalty, when the regularization parameter is tuned. The elastic net is formulated as the optimization problem
$$
\underset{\beta\in\R^p}{\min} \|Y-\bb X\beta\|_2^2 + \lambda\left[ \frac{1}{2}(1-\alpha)\|\beta\|_2^2 + \alpha\|\beta\|_1 \right]
$$
which is a strictly convex program for parameters $\lambda>0$, $0< \alpha< 1$. Therefore, for solving the optimization problem even traditional numerical methods are effective, e.g. the block coordinate descent that subsequently minimize the objective function cyclically following suitable directions spanned by one or more coordinate axes with a step-size controlled by some line search \cite{nocedalwright2006}.

The class of \textit{matching pursuit} algorithms, based on the greedy search in the frame for additional vectors which are maximally coherent to the residual representation error, contains the well-known Orthogonal Matching Pursuit (OMP) that is a prominent representative for being simple as well as reasonably effective. The statistical counterpart corresponds to an approach similar to the forward stepwise regression \index{forward stepwise regression} procedure \cite{draper1998applied} with only one variable LS fit for the residual. This class of variable selection-based regression procedures are well-known since the 1960s but suffer from yielding a highly suboptimal subset of explanatory variables in facts and erroneous inferences due to the multiple hypothesis testing problem, that is traditionally dealt with using Bonferroni-type procedures \cite{benjamini1995controlling}. Another partial remedy to this issue especially in high-dimensional problems is provided by some upstream dimensionality reduction technique.

The OMP \cite{pati1993} method attains an approximate solution to the \ref{eq:SAP} problem in the following manner: it starts with setting $\beta=0$ and selecting the column $\bb X_j$ of $\bb X$ minimizing the residual $r^{(1)} = \| Y - \bb X_j \beta_j\|_2$ w.r.t. to the $j$-th coefficient $\beta_j$. Afterward, it adds another column $\bb X_{j'}$ to the selection so that the second residual $r^{(2)} = \| Y - \bb X_j \beta_j - \bb X_{j'} \beta_{j'} \|_2$ is minimized w.r.t to $\beta_{j'}$ and then orthogonally projects $Y$ onto the span of the updated selection $\{\bb X_j, \bb X_{j'}\}$ so to re-tune $\beta_j$ and $\beta_{j'}$. Cycling $s$ times through these two steps of vector selection and orthogonal projection yields a pool $S\subseteq\{1,...,p\}$ of $s$ column indices and the corresponding residual
$$
r^{(s)} = \| Y - \sum_{j\in S}\bb X_j \beta_j\|_2 \qquad  ,|S|=s,
$$
which is taken as current solution. The iteration is repeated augmenting the pool $S$ with new atom indices until meeting a stopping criterion, such as reaching the constraint for the residual error or the $\beta$ estimator's sparsity. The method was widely studied and admits some enhanced versions, such as LS-OMP, based on projection onto pooled columns and calculating least squares solutions.

The Least-Squares OMP (LS-OMP) algorithm presented in \cite[p. 38]{elad2010}, which is exactly the one widely known in statistical literature as forward stepwise regression \cite{hastie2009elements}, is sometimes confused \cite{kaur2014dissimilarity} with OMP as stated in the historical explanation work \cite{blumensath2007difference}. The key difference lies in the variable-selection criterion used: while OMP, similarly to MP, finds the predictor variable most correlated with the current residual (i.e., performs the single-variable OLS fit), LS-OMP searches for a predictor that best improves the overall fit, that is, solves the full OLS problem on the current support inclusive of the candidate variable. Though this step is more computationally expensive than the single-variable fit, few optimized implementations are available making it more efficient \cite{elad2010,hastie2009elements}. Subsequently to variable selection, all entries in the current support are updated, so the solution and residual recomputing step of LS-OMP coincides with that of OMP.

Another computationally efficient variant of OMP for large samples is based on batch sparse-coding, and is known as Batch-OMP algorithm \cite{rubinstein2008}: it considers pre-computations to reduce the total amount of work involved in coding the entire set of vectors $Y$, and at each iteration the atom selection phase avoids explicitly computing the residual vector $r^{(s)}$ and the projection $\beta_S = \bb X_S^\dagger Y$, but requires knowing only $\bb X^T r^{(s)}$. Other several numerically optimized implementations of OMP using QR and Cholesky decompositions can be found in \cite{sturm2012comparison} with their complexity assessment.

A further class of sparse estimation methods relies on the relaxation of the $\ell_0$-norm by means of smoother functionals approximating $\ell_0$ that promote the sparsity of the solution vector $\beta$, for instance the (pseudo)norms $\|\beta\|_q=(\sum_i |\beta_i|^q)^{1/q}$, $0<q<1$. An interesting example hereof is FOCUSS, namely the FOCal Underdetermined System Solver \cite{gorodnitsky1997sparse}, for it exploits a well-devised optimization technique called iteratively reweighted least squares (IRLS) \cite{green1984}, that is based on the observation \cite{elad2010} that $\|a\|_q^q = \|A^{-1} a\|_2^2$ for an invertible matrix $A = \diag\{|a_i|^t\}_i$ when choosing $t = 1-q/2$. Hence, from a current iterate $\beta^k$, the algorithm computes the next iterate $\beta^{k+1}$ as solution to the weighted least squares problem (WLS)
$$
\underset{\beta\in\R^p}{\min} \|B_k^+ \beta\|_2^2 \qquad \text{subject to } Y = \bb X\beta
$$
where $B_k=\diag\{|\beta^k_i|^t\}_i$ and $B_k^+$ denotes its Moore-Penrose pseudoinverse.
%We will see that the idea of formulating a weighted least squares subproblem is used in the second phase of our algorithm as well.
Despite the fact that FOCUSS heuristic does not guarantee the attaining of a local minimum point of the $\ell_q$ relaxed problem, it converges to some fixed point and has the nice property of stabilizing a coefficient of the partial solution $\beta^k$ as soon as it becomes zero during the iterations, thus promoting the sparsity \cite[\S 3.2.1]{elad2010}. The method yields a sequence of iterates converging to limit points that are minima of the descent function $L(\beta)=\Pi_{i} |\beta_i|$ \cite{gorodnitsky1997sparse}.

Another method of $\ell_0$-norm approximation called L0ADRIDGE \cite{liu2017sparse} was proposed for feature selection and prediction tasks in sparse generalized linear models with big omics data. The method formulates the sparse estimation problem as a maximum likelihood problem
$$
\underset{\beta\in\R^p}{\argmin} -\mathcal L(\beta) + \lambda\|\beta\|_0
$$
with the $\ell_0$ penalization term which is then suitably approximated introducing an auxiliary variable $\eta$ replacing the $\beta$ in the penalization term and shadowing the original $\beta$ in the iterations of the unconstrained optimization process: such process is carried out for all variables but $\eta$ using standard Newton-Raphson iterations, and the vector $\eta$ is reassigned $\beta$ at the end of each iteration.
%One major difference with our algorithm is that we do not formulate a problem with penalization term nor apply any approximation adding auxiliary variables.
The L0ADRIDGE method performed well on sparse regression for suboptimal debulking prediction in ovarian cancer data \cite{liu2017sparse}.

%In the remaining subsections we describe analytical details of our estimation strategy.

%[*** PER VISIONE BAYESIANA: LIBRO RISH SEZ. 2.8 ***]

\subsection{Bayesian interpretation}
A modern view is given by the Bayesian interpretation \cite[\S 2.8]{rish2014sparse} of the regularization term-constrained loss minimization problem \eqref{prob:constrained_minloss}. Such problem can be reformulated introducing Lagrange multiplier $\lambda$ as
\begin{equation}\label{prob:min_regularizedloss}
\min_\beta\ L(\beta; Z, \mathbb D) + \lambda J(\beta).
\end{equation}
Suppose that the data are distributed with a probability $p(Z,\mathbb D \mid \beta)$ and, adhering to Bayesian approach, the parameter $\beta$ follows a prior distribution $p(\beta | \lambda)$ governed by the hyperparameter $\lambda$. The method of maximum a posterior (MAP) estimation in Bayesian statistics yields the estimator $\hat\beta_{\text{MAP}}$ that turns out to be the maximizer of the joint probability $p(\beta,Z,\mathbb D) = p(Z,\mathbb D | \beta)p(\beta|\lambda)$. Taking the negative logarithm one obtains $-\log p(Z,\mathbb D | \beta) -\log p(\beta|\lambda)$, allowing to formulate the equivalent MAP problem
$$
\min_\beta -\log p(Z,\mathbb D | \beta) -\log p(\beta|\lambda).
$$
The first term, which is the negative log-likelihood, takes the role of the loss function $L(\beta; Z,\mathbb D)=-\log p(Z,\mathbb D | \beta)$, while the second term, which is a function of the prior probability $p(\beta|\lambda)$ on the parameter, is a function of the kind $R(\beta,\lambda)$, which takes the form of $R(\beta,\lambda) = \lambda J(\beta)$ in the Lagrange multiplier formulation \eqref{prob:min_regularizedloss}. The Bayesian view hence interprets the regularized maximum likelihood estimation for $\beta$ with regularization control parameter $\lambda$ as MAP estimation with hyperparameter $\lambda$ for the prior on $\beta$.

The interpretation can be evidenced concretely in the noteworthy case of $\ell_1$ regularized least squares loss problem. Indeed, assume a linear model, where the response variables $Y_i$ are i.i.d. with Gaussian distribution $\mathcal N(\mathbb X_i \beta, 1)$, having denoted with $\mathbb X_i$ the $i$-th row of data matrix $\mathbb X$, namely they have conditional PDF
$$
p(y_i|\mathbb X_i \beta) = \frac{1}{\sqrt{2\pi}} e^{-\frac{1}{2}(y_i-\mathbb X_i \beta)^2}.
$$
The negative logarithm of the likelihood function $p(Y,\mathbb X\mid \beta) = p(Y|\mathbb X\beta)p(\mathbb X)$ can be expressed by direct calculation
$$
L(\beta; Y,\mathbb X) = \frac{1}{2} \|Y-\mathbb X \beta\|_2^2 + c
$$
where $c$ is a constant, which can be ignored for optimization goals. We can recast the problem in Bayesian statistics, assuming that the $\beta_i$'s are i.i.d. having Laplace prior distribution with hyperparameter $\lambda$:
$$
p(\beta_i|\lambda) = \frac{\lambda}{2} e^{-\lambda|\beta_i|}.
$$
It is straight-forward to see that the MAP estimation turns out to be formulated as the optimization problem:
$$
\min_\beta\ \|Y-\mathbb X \beta\|_2^2 + 2\lambda\sum_i |\beta_i|
$$
where the first term is the squares loss function $L(\beta; Y,\mathbb X)= \|Y-\mathbb X \beta\|_2^2$ and the second term is the $\ell_1$ regularizer $R(\beta,\lambda) = 2\lambda\|\beta\|_1$. Therefore, the Bayesian treatment of the linear Gaussian observations model with a Laplace prior yields a MAP estimator that corresponds to the Lagrange multiplier formulation of the \ref{prob:Lasso} problem.
Beyond the Lasso formulation, other references to statistical models with loss functions and regularization terms promoting parameter's sparsity can be found in \cite{rish2014sparse}.

\section{Sparse recovery conditions}
\subsection{Null Space Property and Spark}
Despite the sparse optimization problem \eqref{PJ} enjoys different properties for the cases of hard sparsity and convex variant, namely for $J$ being $\ell_0$ and $\ell_1$ norms, their solutions coincide in certain cases.
Indeed, in this section we introduce the conditions for ensuring that the unique solution of (\ref{P1}) is also the solution of (\ref{P0}). In this regard, given $z \in \R^m$ and $\Lambda \subset \{ 1, 2, \dots, m\}$, we denote by $z_\Lambda\in\R^m$ the vector with entries
\[
(z_\Lambda)_i = \begin{cases} 
	z_i, & i \in \Lambda \\  
	0, & i \not\in	 \Lambda.
\end{cases} 
\]
Sometimes, with a little abuse of notation, the vector of $\R^{|\Lambda|}$ obtained from $z_\Lambda$ by erasing the entries at positions off $\Lambda$ will be again denoted with $z_\Lambda$, when unambiguous from the context.

\begin{definition}
A matrix $\Phi \in \R^{n\times m}$ has the \textit{Null Space Property}\footnote{A term coined by Cohen et al. \cite{cohen2009compressed}.} (NSP)  of order $k$ with constant $\gamma>0$, for any\footnote{In this chapter, we assume the standard bases of $\R^n$ and $\R^m$, and hence consider a linear map $\R^m\rightarrow\R^n$ and its representation matrix $\Phi \in \R^{n\times m}$ w.r.t. the standard bases as the same, so we can write the null space of such linear map as $\ker\Phi$.} $z \in \ker\Phi$ and $\Lambda \subset  \{1, 2, \dots, m\}$, $|\Lambda| \leq k$, it holds
\begin{equation}\label{NSP}
\| z\|_p \leq \gamma \| z_{\Lambda^c}\|_p.
\end{equation}
\end{definition}
Notice that the last inequality in the NSP directly implies
$$
\|z_\Lambda\|_p\leq \gamma \| z_{\Lambda^c}\|_p.
$$
Also, a weaker form could be given restating the inequality as $\|z_\Lambda\|_1 < \| z_{\Lambda^c}\|_1$ for all $z\in\ker\Phi \smallsetminus \{0\}$.
The NSP captures the condition that the vectors in the kernel of $\Phi$ shall have non-zero entries that are not too much concentrated on few positions. Indeed, if $z\in\ker\Phi$ is $k$-sparse, then $\|z_{\Lambda^c}\|_1=0$ for $\Lambda=\supp(z)$. The NSP would imply $z_\Lambda=0$ as well. This means that, for matrices $\Phi$ enjoying the NSP of order $k$, the only vector $z\in\ker\Phi$ that is $k$-sparse is $z=0$.

Since in general the solutions to $\eqref{P1}$ does not coincide with the solutions to $\eqref{P0}$, the hope is to find some cases where the solutions are the same. The Null Space Property provides precisely necessary and sufficient conditions \cite{Gribonval2003,Donoho2005hds,Rauhut10} for solving the problem (\ref{P1}). Indeed, we have:
\begin{theorem}
Given a matrix $\Phi\in\R^{n\times m}$ , a $k$-sparse vector $x\in\R^m$ is the unique solution of \eqref{P1} with $s=\Phi x$ if and only if $\Phi$ satisfies the NSP of order $k$.
\end{theorem}
This results not only concerns the \ref{P1} problem, but it gives also the solution to $\eqref{P0}$ through the minimization in $\eqref{P1}$. This means that, as direct consequence, if a sensing matrix $\Phi$ has the Null Space Property \index{Null Space!property} of order $k$ it is guaranteed that the unique solution of (\ref{P1}) is also the solution of (\ref{P0}) when it is $k$-sparse. Indeed, if $\hat\alpha$ is a minimizer of the \ref{P0} problem with $s=\Phi x$, then $\|\hat\alpha\|_0\leq \|x\|_0$, so $\hat\alpha$ is $k$-sparse as well. Since it is $k$-sparse, it must be the unique solution $\hat\alpha=x$ in the theorem.

If $\Phi$ has the Null Space Property, the unique minimizer of the (\ref{P1}) problem is recovered by the basis pursuit (BP) algorithm.
Notice that assessing the Null Space Property of a sensing matrix is not an easy task: checking each point in the null space with a support less than $k$ would be prohibitive. Indeed, deciding whether a given matrix has the NSP is NP-hard and, in particular, so is it to compute the relative NSP constant $\gamma$ for a given matrix and order $k>0$ \cite{tillmann2013computational}, but nonetheless it conveys a nice geometric characterization of the exact sparse recovery problem.

Another linear algebra tool which is useful for studying the sparse solutions is related to the column spaces of a matrix. We know that the column rank \index{rank!column} of a matrix $\Phi$ is the maximum number of linearly independent column vectors of $\Phi$. Equivalently, the column rank of $\Phi$ is the dimension of the column space of $\Phi$.
A criteria to assess the existence of a unique sparsest solution to a linear system is based on the notion called spark\cite{Donoho2003} \index{spark} of a matrix defined as follows.
\begin{definition}
Given a matrix $\Phi$, $spark(\Phi)$ \index{spark} is the smallest number $s$ such that there exists a set of $s$ columns in $\Phi$ which are linearly dependent:
$$
spark(\Phi) = \min \{\|z\|_0 : \Phi z = 0, z\neq 0\}.
$$
\end{definition}
Namely, it is the minimum number of linearly dependent columns of $\Phi$, or equivalently the least sparsity of a non-trivial vector of $\Phi$'s kernel. The spark of a matrix is strictly related to the Kruskal's rank, denoted $krank(\Phi)$, that differs from the well-known (Sylvester) rank and is defined as the maximum number $k$ for which every subset of $k$ columns of the matrix $\Phi$ is linearly independent; of course $krank(\Phi)\leq rank(\Phi)$. So in these terms, we have that $2\leq spark(\Phi)=krank(\Phi)+1\leq rank(\Phi)+1$. Typically, the last inequality turns into an equality: for instance it happens with probability 1 when the matrix $\Phi$ has i.i.d. entries from a Gaussian distribution.

Notice that by definition of spark, we can see from another viewpoint that every non-zero vector $z\in\ker\Phi$ has $\|z\|_0\geq spark(\Phi)$ since it is necessary to linearly combine at least $spark(\Phi)$ columns of $\Phi$ to form the zero vector.
\begin{theorem}
\cite{foucart2013mathematical}
Given a linear system $\Phi\alpha = s$, any $k$-sparse vector $\alpha\in\R^m$ is the unique solution of the system if and only if $krank(\Phi) \geq 2k$.
\end{theorem}
The conditions consists in having every set of $2k$ columns of $\Phi$ being linearly independent.
The spark is a major tool since it provides a simple criterion for the uniqueness of sparse solutions in a linear system. Indeed, using the spark we can easily show:
\begin{theorem}\label{th:spark_sparsest}
\cite{Donoho2003} Given a linear system $\Phi \alpha = s$, if $\alpha$ is a solution satisfying 
$$
\| \alpha \|_0 < \frac{spark(\Phi)}{2}
$$
then  $\alpha$ is also the unique sparsest solution.
\end{theorem}
\begin{proof}
Let $\beta$ be another solution of the linear system, and $\| \beta \|_0 \leq \| \alpha \|_0$. This implies that $\Phi (\alpha - \beta) = 0$. By definition of spark
\begin{equation}\label{eq:spark}
\| \alpha \|_0 + \| \beta \|_0 \geq \| \alpha - \beta\|_0 \geq spark(\Phi).
\end{equation}
Since $\| \alpha\|_0 < \frac{spark(\Phi)}{2}$, it follows that $\| \beta \|_0 \leq \| \alpha \|_0 < \frac{spark(\Phi)}{2}$. 
By eq. \eqref{eq:spark}
\[
spark(\Phi) \leq \| \alpha \|_0 + \| \beta \|_0 <  \frac{spark(\Phi)}{2} + \frac{spark(\Phi)}{2}  = spark(\Phi)
\] 
which yields a contradiction.
\end{proof}
While computing the rank of a matrix is an easy task, from a computational point of view, the problem of computing the spark is difficult. In fact, it has been proved to be an NP-hard problem \cite{tillmann2013computational}. This difficulty motivates the need for a simpler way to guarantee the uniqueness, as we are going to outline in the next sections through other geometric tools.

\subsection {Restricted Isometry Property}
Compressive sensing allows to recover sparse signals accurately from a very limited number of measurements, possibly contaminated with noise, relying on the properties of the sensing matrix, such as the Restricted Isometry Property (RIP). A nice feature of such condition is that it usually holds for commonly used random matrices, such as those with i.i.d. entries drawn from many families of probability distributions. The RIP is predominantly used to establish performance guarantees when either the measurement vector $s$ is corrupted with noise or the vector $\alpha$ is not strictly $k$-sparse \cite{Tanner2011}. This stability feature is essential for practical algorithms since the measurements are rarely free from noise in applications.

The previously introduced Null Space Property is a necessary and sufficient condition to ensure that any $k$-sparse solution vector $\alpha$ is recovered as the unique minimizer of the problem (\ref{P1}).
When the signal $s$ is contamined by noise it will be useful to consider stronger condition like the Restricted Isometry Property \index{Restricted Isometry!property} condition on matrix $\Phi$, introduced by Candès and Tao \cite{Candes2005}, and defined as follows.
\begin{definition}
A matrix $\Phi$ satisfies the Restricted Isometry Property (RIP) of order $k$ if there exists a constant $\delta_k \geq 0$ such that 
\begin{equation}
\label{RIP}
(1-\delta_k) \|\alpha\|_2^2 \leq \| \Phi \alpha \|_2^2 \leq (1+\delta_k)\|\alpha\|_2^2 
\end{equation}
holds for all $\alpha \in \Sigma_k$. The smallest of these constants $\delta_k$ is called the Restricted Isometry Constant (RIC).
\end{definition}
If a matrix $\Phi$  satisfies the RIP of order $2 k$, then we can interpret eq. (\ref{RIP}) as saying that $\Phi$ approximately preserves the distance between any pair of $k$-sparse vectors $x,y$, simply setting $\alpha = x-y\in\Sigma_k$.
That is to say, multiplying by every subset of at most $k$ columns of $\Phi$ behaves very close to an isometric transformation, where the relative closeness is expressed in terms of the RIP constant $\delta_k$.
If the matrix $\Phi$ satisfies the RIP of order $k$ with constant $\delta_k$, then for any $k' < k$ we automatically have that $\Phi$ satisfies the RIP of order $k'$ with constant  $\delta_{k'} \leq \delta_{k}$. This monotonicity is one of the main properties of the RIC described in the following results. Remind that given an operator $T:U\rightarrow V$ between vector spaces $U$ and $V$, endowed with norms $\|\cdot\|_U$ and $\|\cdot\|_V$ respectively, the operator norm of $T$ is $\|T\|_{op}:=\inf\{c\geq 0:\|Tx\|_V\leq c\|x\|_U$ for all $x\in U\} = \sup\{\|Tx\|_V/\|x\|_U :x\neq 0\}$, and in particular for matrices $T$ the operator norm of $T$ is the largest singular value $\sigma_1(T)$ of $T$.
\begin{proposition}
Let the matrix $A\in\R^{n\times m}$ satisfy the RIP with RICs $\delta_k$, for orders $k=1,2, ...$. Then
\begin{enumerate}[label=(\roman*)]
\item The sequence of RICs $\{\delta_k\}$ is non-decreasing, i.e. $\delta_1\leq\delta_2\leq \cdots \leq \delta_m$
\item The restricted isometry constant $\delta_k$ can be evaluated equivalently as the maximal $\ell_2$-norm distortion on $k$-sparse vectors:
$$
\delta_k = \max_{\Lambda\subset [N]:|\Lambda|\leq k} \|A_\Lambda^T A_\Lambda -I_k\|_{op}
$$
\end{enumerate}
\end{proposition}
Notice that, by definition of operator norm, the last equality is
\begin{align*}
\delta_k &= \sup_{\Lambda\subset [N]:|\Lambda|\leq k,\ x\in\R^k,x\neq 0} \frac{\|A_\Lambda^T A_\Lambda x -I_k x\|_2}{\|x\|_2}
= \sup_{x\in\R^m:\|x\|_0\leq k} \frac{\|A^T A x -I_m x\|_2}{\|x\|_2} =\\
& = \sup_{x\in\R^m:\|x\|_0\leq k} \frac{|x^T A^T A x - x^T x|}{x^T x}
\end{align*}
That is, $\left|\|Ax\|_2^2 - \|x\|_2^2\right|\leq \delta_k \|x\|_2^2$ when $\|x\|_0\leq k$, which is indeed equivalent to the RIP with constant $\delta_k$.

For matrices $\Phi$ satisfying RIP the RIC can be calculated \cite{foucart2013mathematical} in practical terms from the smallest and largest singular values of any subset $\Lambda$ of $k$ columns of $\Phi$:
$$
\delta_k = \max_{i, |\Lambda|\leq k} |\sigma_i(\Phi_\Lambda) - 1|
= \max\{ \max_{|\Lambda|\leq k} |\sigma_1(\Phi_\Lambda) - 1|, \max_{|\Lambda|\leq k} |\sigma_n(\Phi_\Lambda) - 1| \}.
$$
In other words, all singular values of submatrices $\Phi_\Lambda$, for $|\Lambda|\leq k$, are in the interval $[1-\delta_k, 1+\delta_k]$. When $\delta_k<1$ the left-hand side of RIP's inequality ensures that $\ker\Phi_\Lambda =\{0\}$, namely it is injective, so usually the condition $\delta_k\in (0,1)$ is replaced in the definition. Actually, for $k$-sparse vectors the condition $\delta_{2k}<1$ is more interesting since it yields $\Phi(\alpha-\beta)\neq 0$ for $\alpha\neq\beta$, so distinct $k$-sparse vectors have distinct measurement vectors, which guarantees recoverability.

Finally, for completeness, we highlight the relationship between the RIP and the mutual coherence\index{mutual coherence} $\mu(\Phi)$, as well as the RIP versus the Nullspace Property \cite{foucart2013mathematical,rish2014sparse}.
\begin{proposition}
Let $\Phi$ be a matrix with unit $\ell_2$-norm columns. Then RIC satisfies:
\begin{enumerate}[label=(\roman*)]
\item $\delta_1=0,\ \delta_2=\mu(\Phi)$
\item $\delta_k\leq (k-1)\mu(\Phi)$
\end{enumerate}
\end{proposition}
\begin{proposition}
Let $\Phi$ have the RIP of order $2k$ with RIC $\delta_{2k}<\sqrt{2}-1$. Then $\Phi$ satisfies the NSP of order $2k$ with constant
$$
\gamma = \frac{\sqrt{2}\delta_{2k}}{1-(1+\sqrt{2}) \delta_{2k}}
$$
\end{proposition}
The former result provides bounds to the restricted isometry constant in terms of the mutual coherence, while the latter shows that if a matrix satisfies the RIP, then it also satisfes the NSP. Thus, the RIP is a condition stronger than the NSP.

The RIP can be also described by the effect of the matrix $\Phi$ on the norm of the vectors, bounding the rate of change for the function defined as $f(\alpha) = \|\Phi\alpha\|_2^2$. The continuously differentiable functions $f:\R^m\rightarrow\R$ satisfying the condition
$$
\frac{a}{2}\|x-y\|_2^2\leq f(y) - f(x) - \langle \nabla f(x), y-x \rangle \leq \frac{b}{2}\|x-y\|_2^2 \quad \text{for all } x,y\in C\subseteq \R^m
$$
are said to be $a$-Restricted Strong Convex (first inequality) and $b$-Restricted Strong Smooth (second inequality). These inequalities correspond to classical convexity and smoothness conditions on differentiable functions simply restricted to a region $C$ that could be even non-convex. The RIP of constant $\delta_k$ of a matrix $\Phi$, for even integer $k>1$, can be characterized by this condition noticing that, taking the function $f(\alpha) = \|\Phi\alpha\|_2^2$, it can be straight-forward to check that the convexity/smoothness constants can be set to $a=2-2\delta_k$ and $b=2+2\delta_k$ when restricting to $k/2$-sparse vectors, $C=\Sigma_{k/2}$.

It is of interest to understand the dependence between the number of observations $n$, i.e. rows of the sensing matrix $\Phi$, and the desired RIC $\delta_k$. In order to quantify this dependence, one can exploit results regarding suitably designed matrices, and in particular the Johnson-Lindenstrauss lemma, which concerns the embedding of finite sets of points in low-dimensional spaces \cite{johnson84}. The Johnson-Lindenstrauss lemma is not inherently connected with sparsity per se, but it can lead to RIP for certain matrices.
\begin{theorem}[Johnson-Lindenstrauss Lemma \cite{johnson84}]
Let $X\subset \R^m$ be a set of $N=|X|$ points and let $0<\varepsilon<1/2$ be arbitrary. Then there exists a map $T:\R^m\rightarrow\R^n$ for some $n=O(\varepsilon^{-2}\log N)$ such that
\begin{equation}\label{eq:johnson-lindenstrauss}
(1-\varepsilon) \|\alpha-\beta\|_2^2 \leq \|T(\a)-T(\beta)\|_2^2 \leq (1+\varepsilon)\|\a-\beta\|_2^2
\end{equation}
for every $\a,\beta\in X$.
\end{theorem}
In \cite{larsen2017optimality} it is also shown that, when $\varepsilon > 1/(\min\{N,m\})^{0.4999}$ a set $X$ requiring the low dimension estimate $\Omega(\varepsilon^{-2}\log N)$ can be effectively constructed, therefore $n=\theta(\varepsilon^{-2}\log N)$ is actually the optimal estimate for having the concentration inequality \eqref{eq:johnson-lindenstrauss}.

In compressive sensing, \index{compressive sensing} random matrices are usually applied as random projections of a high-dimensional space with sparse or compressible signal vectors onto a lower-dimensional space that with high probability contains enough information to enable exact or small error signal reconstruction.
\begin{theorem}[Distributional Johnson-Lindenstrauss Lemma \cite{johnson84}]
For any dimension $m\in\N_+$ and $\varepsilon,\delta\in(0,1)$ there exists a probability distribution $\mathcal D$ over all linear mappings $T:\R^m\rightarrow\R^n$, where $n=\theta(\varepsilon^{-2}\log \frac{1}{\delta})$ such that
$$
\mathbf P\left(\left| \|T(\a)\|_2^2 - \|\a\|_2^2 \right|\leq \varepsilon\|\a\|_2^2 \right) \geq 1-\delta \qquad \text{for all }\a\in\R^m
$$
where $T$ has probability distribution $\mathcal D$.
\end{theorem}
Random sensing matrices $\Phi$ drawn according to any distribution that satisfies the Johnson-Lindenstrauss concentration inequality \cite{johnson84} have been shown to satisfy the Restricted Isometry Property with high probability \cite{DeVore08,rish2014sparse}.
\begin{proposition}
Let $\Phi$, be a random matrix \index{matrix!random} of size $n\times m$ drawn according to any distribution that satisfies the concentration inequality 
\[
\mathbf{P} \left( \big| \| \Phi \alpha \|_2 - \| \alpha \|_2 \big| \geq \epsilon \| \alpha \|_2 \right) \leq 2 e^{-n c_0(\epsilon)}, \qquad \text{for\ } 0 < \epsilon < 1
\] 
where $c_0(\epsilon) > 0$ is a function of $\epsilon$.\\  
Then for any $0 < \delta < 1$, we have that for all $\alpha \in \Sigma_k$, $k<n$:
\[
(1-\delta) \| \alpha \|_2^2 \leq \| \Phi \alpha \|_2^2 \leq (1+\delta) \| \alpha \|_2^2 
\]
holds with a probability at least
$$
1-2(9/\delta)^ke^{-n c_0(\delta/2)}
$$
that is, the RIP of order $k$ and constant $\delta$ holds with the stated probability lower bound.
\end{proposition}
When $\Phi \sim N(0, \frac{1}{n}I)$, one can take as $c_0$ the monotonically increasing function $c_0 = \frac{\epsilon^2}{4} - \frac{\epsilon^3}{6}$. Unfortunately, if $\Phi$ has a large number $m$ of columns, estimating and assessing the Restricted Isometry Constant is computationally impractical. A computationally efficient, yet conservative, estimate for ensuring the Restricted Isometry Property can be obtained through the mutual coherence. To this aim, in the next section we introduce some bounds for the mutual coherence of a dictionary $\Phi$.

\subsection {Mutual Coherence}
Conditions on the mutual coherence\index{coherence!mutual} can lead to the uniqueness and recoverability of the sparsest solution. While computing Restricted Isometry Property, Null Space Property and spark are NP-hard problems, the coherence of a matrix can be evaluated more effectively.
\begin{definition}
Let $\phi_1, \dots, \phi_m$ the columns of the matrix $\Phi$. The mutual coherence of $\Phi$ is then defined as
\[ 
	\mu(\Phi) = \max_{i<j} \frac{| \phi_i^{T} \phi_j |}{\| \phi_i \|_2 \| \phi_j \|_2}.
\]
\end{definition}
Mutual coherence is also known as maximal frame correlation. This is in fact the largest modulus of the cosine between two vectors in the dictionary $\Phi$, i.e. the maximum absolute cosine similarity.

By Schwartz inequality, $0 \leq \mu(\Phi) \leq 1$. We say that a matrix $\Phi$ is incoherent if $\mu(\Phi)=0$.
For  $n \times n$ unitary matrices, columns are pairwise orthogonal, so the mutual coherence is obviously zero. For full rank $n \times m$ matrices $\Phi$ with $m>n$, $\mu(\Phi)$ is strictly positive, and it is possible to show \cite{Strohmer03} that the following inequality, called Welch bound, holds:
\[
	\mu (\Phi) \geq \sqrt{\frac{m-n}{n(m-1)}}
\]
with the equality attained only for a family of matrices in $\R^n$ named, by definition, optimal Grassmanian frames\index{frame!Grassmanian}. 
Moreover, if $\Phi$ is a Grassmanian frame, the $spark(\Phi) = n + 1$, the highest value possible.\\

Mutual coherence is easy to compute and give a lower bound to the spark. In order to outline this result, we briefly recall the Gershgorin's Theorem for localizing eigenvalues of a matrix, which is extensively used for perturbation methods in applied mathematics \cite[\S 6]{horn2012matrix}. Given a $n \times n$ matrix $A = \{ a_{i,j} \}$, let be $R_k = \sum_{j \neq k} |a_{k,j}|$. The complex disk $D_k =\{ z:| z - a_{k,k} |\leq R_k \}$ is called a Gershgorin's disk\index{Gershgorin!theorem}, $1 \leq k \leq n$. The Gershgorin's Theorem \cite{Gershgorin1931} states that every eigenvalue of $A$ belongs to (at least) one Gershgorin's disk. The theorem is a commonly used tool for delimiting estimated regions for the eigenvalues and related bounds simply on the basis of matrix entries.

\begin{theorem}
 \cite{Donoho2003} For any matrix $\Phi \in \R^{n \times m}$ the spark of the matrix is bounded by a function of its mutual coherence as follows:
 \[
   spark(\Phi) \geq 1 + \frac{1}{\mu(\Phi)}.
 \]
\end{theorem}
\begin{proof}
Since normalizing the columns does not change the coherence of a matrix, without loss of generality we consider each column of the matrix $\Phi$ normalized to the unit $\ell_2$-norm. Let $G = \Phi^T \Phi$ the \index{matrix!Gram}Gram matrix of $\Phi$. Consider an arbitrary minor from $G$ of size $p \times p$, built by choosing a subset of $p$ columns from the matrix $\Phi$ and computing their relative sub-Gram matrix $M$. We have $|\phi_i^{T} \phi_j | = 1$ if $k = j$ and $|\phi_i^{T} \phi_j | \leq \mu(\Phi)$ if $k = j$, as consequence $R_k \leq (p-1)\mu(\Phi)$.

It follows that Gershgorin's disks are contained in $\{z : |1 - z| \leq (p-1) \mu(\Phi) \}$. If $(p-1)\mu(\Phi) < 1$, by Gershgorin's theorem, $0$ cannot be eigenvalues of $M$, hence every $p$-subset of columns of $\Phi$ is composed by linearly independent vectors. We conclude that a subset of columns of $\Phi$ linearly dependent should contain $p \geq 1 + \frac{1}{\mu(\Phi)}$ elements, hence $spark(\Phi) \geq 1 + \frac{1}{\mu(\Phi)}$.
\end{proof}

The previous result together with Theorem \ref{th:spark_sparsest} leads to the following straight-forward condition implying the uniqueness of the sparsest solution in a linear system $\Phi \alpha = s$.
\begin{theorem} 
\label{th:unique_coherence}
\cite{Donoho2003} If a linear system $\Phi \alpha = s$ has a solution $\alpha$ such that
$$
\|\alpha\|_0 < \frac{1}{2}\left[1+\frac{1}{\mu(\Phi)}\right]
$$
then $\alpha$ is also the unique sparsest solution.
\end{theorem}
Notice that the mutual coherence can never be smaller than $\frac{1}{\sqrt{n}}$ and therefore the sparsity bound of Theorem \ref{th:unique_coherence} cannot be larger than $\frac{\sqrt n}{2}$. In general, since Theorem \ref{th:spark_sparsest} uses the spark of the matrix, it gives a sharper and more powerful property than the last theorem, which results to be a rather useful feature in dictionary learning applications, but the latter one entails a lower computational complexity.

The notion of mutual coherence was then later generalized from maximal absolute cosine similarity between a pair of vectors to the maximal total absolute cosine similarity of any group of $p$ atoms with respect to the rest of the dictionary \cite{tropp2004greed}. Although this is more difficult to compute than the mutual coherence, it is a sharper tool.

\section{Algorithms for Sparse Recovery}
The problem we analyze in this section is the approximation of a signal $s$ using a linear combination of $k$ columns of the dictionary $\Phi \in \R^{n \times m}$. In particular we seek a solution of the 
minimization problem
\begin{equation}
 \label{min_mp}
  \min_{\Lambda \subset [m]:|\Lambda|=k} \min_{\alpha_\lambda} \| \sum_{\lambda \in \Lambda} \phi_\lambda \alpha_\lambda - s \|_2^2
\end{equation}
for a fixed $k$ with $1 \leq k \leq m$.
The actual difficulties in solving problem (\ref{min_mp}) stems from the optimal selection of the index set $\Lambda$, since the ``exhaustive search'' algorithm for the optimization requires to test all $\binom{m}{k} \geq \left( \frac{m}{k}\right)^k$ subsets of $k$ columns of $\Phi$; this seems prohibitive for real instances. So remains it if we try to find the sparsest solution $\alpha$ in the noiseless case, i.e. for the linear system $\Phi\a=s$. To show the concrete example in \cite{elad2010}, consider a $500\times 2000$ matrix $\Phi$ and an oracle information stating that the sparsest solution of the linear system has sparsity $k=|\Lambda|=20$. In order to find a corresponding set $\Lambda$ of columns in $\Phi$, one would be tempted to exhaustively sweep through all $\binom{m}{k} = \binom{2000}{20} \approx 3.9\cdot 10^{47}$ choices of the subset $\Lambda$ and test the equality $\Phi_\Lambda \a_\Lambda = s$ for each subset. But even if a computer could perform $10^9$ tests/sec, it would take more than $10^{31}$ years to terminate all tests. This easily motivates the need for devising effective computational techniques for sparse recovery.

The algorithms developed in literature can be grouped into three main classes:
\begin{itemize}
\item \textit{Basis Pursuit methods} \index{Basis Pursuit} where the sparsest solution in the $\ell_1$ sense is desired and there is an underdetermined system of linear equations $\Phi \alpha = s$ that must be satisfied exactly. This is characterized by the fact that the sparsest solution in such sense can be easily solved by classical linear programming algorithms.
\item \textit{Greedy methods} \index{sparse recovery algorithm!greedy} where an approximation of the optimal solution is found by starting from an initial atom and then incrementally constructing a monotone increasing sequence of subdictionaries by locally optimal choices at each iteration.
\item \textit{Convex relaxation methods} that loosen the combinatorial sparsity condition in the recovery problem to a related convex/non-convex programming problem and solve it with iterative methods.
\end{itemize}
We outline some representative algorithms for these classes in this section.

\subsection{Basis Pursuit}
The Basis Pursuit (BP) method seeks the best representation of a signal $s$ by minimizing the $\ell_1$  norm of the coefficients $\alpha$ of the representation. Ideally, we would like that some components of $\alpha$ to be zero or as close to zero as possible.
It can be shown \cite{Rudelson06} that the \ref{P1} problem can be recast into a linear programming problem (LP) in the standard form 
\begin{equation}
	\label{LProb}
	\min_{x \in \R^m} c^T x \mbox{\ \ s.t.\ \ } M x = b, x \geq 0
\end{equation}
where $J(x)=c^T x$ is the objective function, $M x = b$ is a collection of equality constraints and the inequality $x \geq 0 $ is understood element-wise, i.e. a set of bounds.

Indeed, though the objective function of \ref{P1} is not linear but piece-wise linear, we can easily transfer the nonlinearities to the set of constraints by adding new variables $t_1, \dots, t_n$ that turns the original \ref{P1} problem into the following linear programming problem formulation:
\begin{align*}
	\min\hspace{7.5mm} \sum_{i=1}^m t_i \\     								
	\text{s.t.\qquad} \alpha_i -t_i &\leq 0, \quad i=1, \dots, m\\
	-\alpha_i -t_i &\leq 0, \quad i=1, \dots, m\\   										
	\Phi \alpha &= s
\end{align*}
with $2m$ inequalities constraints, that in matrix form are $A(\a,t)^T \leq 0$.
Introducing slack variables $\a'_i$ and $t'_i$, and replacing the variables $\a =\a^+-\a^-$ and $t=t^+-t^-$ with non-negative variables $\a^+,\a^-,t^+,t^-\geq 0$, one can hence write the \ref{P1} problem in LP standard form
\begin{align*}
\label{PL1}
\tag{ $P_{\ell_1}$ }
\min & \ \sum_{i=1}^m (t_i^+-t_i^-)  \\
\mbox{s.t. }  &\ [A, -A, I] (\a^+, t^+,\a^-, t^-, \a', t')^T = 0 \\
&\ [\Phi, 0, -\Phi, 0, 0, 0] (\a^+, t^+,\a^-, t^-, \a', t')^T = s\\
&\ (\a^+, t^+,\a^-, t^-, \a', t')^T \geq 0
\end{align*}
In order to reduce the size of \ref{PL1} problem we can formulate the \textit{dual problem}\index{dual problem}. From duality theory, starting with a linear program in standard form (\ref{LProb}), we can rewrite the problem in the following dual linear program in terms of the dual variables $y$ and $w$ which correspond to the constraints from the primal problem without restrictions
\begin{align*}
\label{DUAL2}
\tag{DLP}
	\min\hspace{0.5mm} & \,s^T y \\
	\mbox{s.t. } & \Phi^T y - 2 w = -e, \mbox{ } 0 \leq v \leq e.
\end{align*}
Once the size of the original problem (\ref{PL1}) was reduced, the dual problem (\ref{DUAL2}) can be solved efficiently by a linear solver \cite{nocedalwright2006}.

%We noticed that BP is known in the statistical literature as Lasso
Moreover, for applications the variant of \ref{P1} problem admitting a measurement error $\varepsilon = \Phi\a - s$ corresponds to the Basis Pursuit Denoising (BPDN) problem \cite{chen1994basis}, which is equivalent to the following Lasso formulation:
$$
\min_{\a\in\R^m} \|\Phi\a - s\|_2^2 + \lambda\|\a\|_1.
$$
Since this is a convex unconstrained optimization problem, there are numerous numerical methods for obtaining one global solution: modern interior-point methods, simplex methods, homotopy methods, coordinate descent, and so on \cite{nocedalwright2006}. These algorithms usually have well-developed implementations to handle Lasso, such as: LARS by Hastie and Efron\footnote{\url{https://cran.r-project.org/web/packages/lars/index.html}}, the $\ell_1$-magic by Candès, Romberg and Tao\footnote{\url{https://candes.su.domains/software/l1magic/}}, the CVX and L1-LS softwares developed by Boyd and students, SparseLab managed by Donoho, SparCo by Friedlander\footnote{\url{https://friedlander.io/software/sparco}}, and SPAMS by Mairal\footnote{\url{http://thoth.inrialpes.fr/people/mairal/spams/}}. For large problems, it is worth to cite the ``in-crowd'' algorithm, a fast method that discovers a sequence of subspaces guaranteed to arrive at the support set of the final global solution of the BPDN problem; the algorithm has demonstrated good empirical performances on both well-conditioned and ill-conditioned large sparse problems \cite{gill2011crowd}.

\subsection{Greedy Algorithms}
Many of the greedy algorithms proposed in literature for carrying out sparse recovery \index{sparse recovery algorithm!greedy} look for a linear expansion of the unknown signal $s$ in terms of functions $\phi_i$: 
\begin{equation}
	\label{eq:rec}
	s = \sum_{i=1}^m \alpha_i \phi_i.
\end{equation}
We may interpret that in such a way the unknown data (signal) $s$ is explained in terms of atoms (functions $\phi_i$) of the dictionary $\Phi$ used for decomposition.
The greedy algorithms for sparse recovery find a sub-optimal solution to the problem of an adaptive approximation of a signal in a redundant set of atoms, namely the dictionary, by incrementally selecting the atoms. In the simplest case, if the dictionary $\Phi$ is an orthonormal basis, the coefficients are given simply by the inner products of the dictionary's atoms $\phi_i$ with the signal $s$, i.e. $\alpha_i = \langle s, \phi_i\rangle$.
However, generally, the dictionary is not an orthonormal basis but redundant. Nonetheless, well-designed dictionaries $\Phi = \{ \phi_i\}_{i=1, \dots, m}$ are those ones properly revealing the intrinsic properties of an unknown signal or, almost equivalently, giving low entropy of the ${\alpha_i}$ and possibilities of good lossy compression.

In applications, the equality condition in \ref{eq:rec}, which in fact corresponds to an exact representation of the signal, is typically relaxed by introducing a noisy model, so that the admitted representation is approximate:
\begin{equation}
\label{eq:rec_approx}
s \approx  \sum_{t=1}^k \alpha_t \phi_{\lambda_t}
\end{equation}
and corresponds to an expansion of $s$ using a certain number, $k$, of dictionary atoms $\phi_{i_t}, t=1,...,k$.

%We may relax the requirement of exact signal representation (), and try to automatically choose the atoms $\phi_{\lambda_t}$, optimal for the representation of a given signal $s$, from a redundant dictionary $\Phi$. The expansion becomes an approximation, and uses only the functions $\phi_{\lambda_t}$ chosen from the redundant dictionary $\Phi$. In practice, the dictionary contains orders of magnitude more candidate functions $\phi_{\lambda_t}$ than the number $k$ of functions chosen for the representation:

A criterion of optimality of a given solution $\a$ based on a fixed dictionary $\Phi$, signal $s$, and certain number $k$ of atoms/functions used in the expansion can be naturally the reconstruction error of the representation
\[
	\epsilon = \| s -  \sum_{t=1}^k \alpha_t \phi_{\lambda_t} \|_2^2
\]
which is a squared Euclidean norm type. As already said, the search for the $k$ atoms of $\Phi$ and the corresponding coefficients is clearly computationally intractable.
%Finding the minimum requires checking all the possible combinations (subsets) of $k$ functions from the dictionary, which leads to a combinatorial explosion. Therefore, the problem is intractable even for moderate dictionary sizes.

The Matching Pursuit (PM) algorithm, proposed in \cite{mallatzhang93}, finds constructively a sub-optimal solution by means of an iterative procedure.
In the first step, the atom $\phi_{\lambda_1}$ which gives the largest magnitude scalar product (interpreted as signal correlation) with the signal $s$ is selected from the dictionary $\Phi$, which is assumed to have unit-norm atoms, i.e. $\|\phi_i\|^2_2 = 1$. At each consecutive step $t>1$, every atom $\phi_i$ is matched with the residual error $r_{t-1}$ calculated subtracting the signal from the approximate expansion using the atoms selected in the previous iterations, that is, after initializing $r_0=s$, it iterates these two steps:
\begin{align*}
\phi_{\lambda_t} &= \argmax_{\phi \in \Phi} | \langle r_{t-1}, \phi\rangle  | \\
r_t &= r_{t-1} - \langle r_{t-1}, \phi_{\lambda_t}\rangle\phi_{\lambda_t}.
\end{align*}
For a complete dictionary, i.e. a dictionary spanning the whole space $\R^n$, the procedure converges, i.e. it produces expansions
$$
\sum_{t=1}^k \langle r_{t-1}, \phi_{\lambda_t}\rangle\phi_{\lambda_t} \rightarrow s
$$
or equivalently $r_t\rightarrow 0$ \cite{mallatzhang93}.
Notice that MP's iteration only requires a single-variable OLS (ordinary least squares) fit to find the next best atom, and a simple update of the current solution and the residual. In such update the residual $r_t$ is not orthogonal with respect to the cumulatively selected atoms, and thus the same atom might be selected again following iterations. Thus, though each iteration of the algorithm is rather simple, the MP (or forward stagewise in statistics literature) may require a potentially large number of iterations for convergence in practice \cite{rish2014sparse}.

Another greedy algorithm, improving the MP, extensively used to find the sparsest solution of the problem (\ref{P0}) is the so-called Orthogonal Matching Pursuit (OMP) algorithm proposed in \cite{Davis94,pati1993} and analyzed by Tropp and Gilbert \cite{Tropp07OMP}.
\begin{algorithm}
	\caption{Orthogonal Matching Pursuit (OMP)}
	\label{alg:OMP}
	\begin{algorithmic}[1]
		\REQUIRE
		\hspace{3.2mm}- a dictionary $\Phi=\{\phi_i\}\in\R^{n\times m}$\\
		\hspace{1cm}- a signal $s\in\R^n$\\
		\hspace{1cm}- a stopping condition\\
		\ENSURE a (sub)optimal solution $\hat\a$ of the \ref{P0} problem with sparsity $\|\hat{\a}\|_0$ equal to the number of iterations determined by the stopping condition
		\vspace{.1cm}
		
		\STATE $r_0 = s, \alpha_0 = 0, \Lambda_0 = \emptyset, t=0$
		\WHILE{ \NOT(stopping condition) } \vspace{.1cm}

		\STATE $\lambda_{t+1} \in \argmax_{j = 1, \dots m} |\langle r_{t}, \phi_j \rangle |$ \qquad\ \ \emph{\small  (fix a tie-breaking rule for multiple \ \ \ maxima cases)}\\[.1cm]
		\STATE $\Lambda_{t+1} = \Lambda_t \cup \left \{ \lambda_{t+1} \right \}$\\[1mm]
		\STATE $\alpha_{t+1} = \argmin_{\beta\in\R^m : \supp(\beta) \subseteq \Lambda_{t+1}} \| \Phi \beta - s \|_2^2$
		\ \ \emph{\small (a full OLS minimization)}\\[.1cm]
		\STATE $ r_{t+1} = s - \Phi \alpha_{t+1}$
		\STATE $ t = t+1$
		\ENDWHILE\vspace{.1cm}
		
		\RETURN $\alpha_t$
	\end{algorithmic}
\end{algorithm}
It differs from MP only in the way the solution and the residual are updated. As can be seen from Algorithm \ref{alg:OMP}, the OMP recomputes the coefficients of all atoms selected in the current support, by solving a full OLS minimization problem over the support augmented with the new atom to be selected, while the MP minimization only involves the coefficient of the most recently selected atom \cite{rish2014sparse}. As result of this operation, OMP (unlike MP) never re-selects the same atom, and the residual vector $r_t$ at every iteration is orthogonal to the current support's atoms, namely selected atoms.
%OMP adds a least squares minimization to each step of the Matching Pursuit.
% Let $\Lambda = \{ \lambda_1, \dots, \lambda_k \}$ list the atoms that have been choosen at step $t$.\\
The $t$-th approximant of $s$ is 
$$
\hat s_t = \Phi\alpha_t = \sum_{j=1}^m \a_{t,j}\phi_j
$$
Despite the OMP update step is more computationally demanding than the MP update, it will consider each variable once only due to the orthogonalization process, thus typically resulting into fewer iterations of the overall loop. The solutions obtained by OMP are more accurate than baseline MP.

A further computational improvement of OMP is the Least-Squares OMP (LS-OMP), whose equivalent statistical counterpart is the so-called forward stepwise regression \cite{hastie2009elements}.  While OMP, similarly to MP, finds the atom of $\Phi$ most correlated with the current residual, i.e. performs an OLS minimization based on single-atom, LS-OMP searches for an atom that improves the overall fit, that is it solves the OLS problem on subspace corresponding to the current support plus the candidate atom. This means that the line 3 is replaced in LS-OMP with
$$
(\lambda_{t+1}, \alpha_{t+1} ) \in \argmax_{j=1,...,m; \a} \|s-\Phi_{\Lambda_t\cup \{j\} }\a \|_2^2.
$$
For this variant of the OMP there are few computationally efficient implementations \cite[p. 38]{elad2010}.

\subsection{Relaxation Algorithms}
An alternative way to solve  \index{sparse recovery algorithm!relaxation} the \ref{P0} problem is to relax its discontinuous $\ell_0$-norm with some continuous or even smooth approximations.
Examples of such relaxation is to replace the $\ell_0$-norm with a convex norm such as the $\ell_1$, some non-convex function like the $\ell_p$-norm for some $p \in ( 0, 1 )$ or other more regular or smooth parametric functions like $f(\a) = \sum_{i=1}^{m}(1-e^{-\lambda \alpha_i^2})$, $f(\a) = \sum_{i=1}^{m}\log(1+\lambda \alpha_i^2)$ or $f(\a) = \sum_{i=1}^{m} \frac{\alpha_i^2}{\lambda + \alpha_i^2}$, for which the parameter $\lambda$ could be tuned for showing analytical properties.

The major hurdles of using $\ell_0$-norm for the optimization stem from its discontinuity and the drawbacks of some combinatorial search.
The main idea of the Smoothed $l_0$ (SL0) algorithm, proposed and analyzed in \cite{Hosein2010,mohimani2008fast}, is to approximate this discontinuous function by a suitable continuous approximant very close to the former, and minimize it by means of optimization algorithms, e.g. steepest descent method. The continuous approximant of $\| \cdot \|_0$ should have a parameter that determines the quality of the approximation.
More specifically, consider the family of single-variable Gaussian functions
\[
 f_\sigma (\alpha) = e^{\frac{-\alpha^2}{2\sigma^2}}
\]
and note that 
\[
 \lim_{\sigma \rightarrow 0} f_{\sigma} (\alpha) =  \begin{cases} 
							  1, & \mbox{if } \alpha=0 \\
							  0, & \mbox{if } \alpha \neq 0.
						      \end{cases}
\]
Defining $F_\sigma (\alpha) = \sum_{i=1}^{m} f_\sigma(\alpha_i)$ for $\a\in\R^m$, it is clear that $F_\sigma \rightarrow \|\cdot\|_0$ pointwise as $\sigma\rightarrow 0$, hence we can approximate $\|\alpha\|_0 \approx m - F_\sigma(\alpha)$ for small values of $\sigma>0$.\\
\begin{algorithm}
\small
  \caption{Smoothed $\ell_0$ (SL0)}
  \label{alg:SL0}
  \begin{algorithmic}[1]
    \REQUIRE
    \hspace{3.8mm}- a dictionary $\Phi\in\R^{n\times m}$ and its Moore-Penrose pseudo inverse $\Phi^\dagger$\\
    \hspace{1cm}- a signal $s\in\R^n$\\
    \hspace{1cm}- a decreasing sequence $\sigma_1, \dots, \sigma_T$\\
    \hspace{1cm}- a stopping condition, a parameter $L$\\
    \ENSURE a feasible point $\hat\a$ of the \ref{P0} that should be close to the optimal solution
    \vspace{.1cm}
   
    \STATE $\alpha_0 = \Phi^\dagger s$, $t=0$
    \WHILE{\NOT(stopping condition)}\vspace{.1cm}
    \STATE $\sigma = \sigma_t$\\[.1cm]
    \STATE Maximize the function $F_\sigma$ over the feasible set $\{\a:\Phi\a=s\}$ using $L$ itera- tions of the steepest ascent algorithm (followed by projection onto the fea- sible set) as follows:
    \FOR {$j = 1,\dots, L$} \vspace{.1cm}
    
      \STATE $\Delta \alpha = \left(\alpha_1 e^{\frac{-|\alpha_1|^2}{2\sigma^2}}, \dots, \alpha_m e^{\frac{-|\alpha_m|^2}{2\sigma^2}} \right)$
      \STATE $\alpha = \alpha - \mu \Delta \alpha$
					\hspace{17mm}\emph{\small (where $\mu$ is a suitable small positive constant)}\\[.1cm]
      \STATE $\alpha = \alpha - \Phi^\dagger (\Phi \alpha - s)$
					\hspace{40mm}\emph{\small (orthogonal projection)}\\[.1cm]

    \ENDFOR
    \STATE $ t = t+1$
    \ENDWHILE
    \RETURN $\a$
   \end{algorithmic}
\end{algorithm}
We can search for the minimum solution in the \ref{P0} problem by maximizing the $F_\sigma(\alpha)$ subject to $\Phi \alpha = s$ for a very small value of $\sigma>0$, which is the parameter that determines how concentrated around 0 the function $F_\sigma$ is. The SL0 method is formalized in Algorithm \ref{alg:SL0}.

The rationale of SL0 is similar to the motivating grounds of those techniques for generating a path of minimizers. Basically, a scheduling of the parameter $\sigma>0$ must be set, producing a decreasing sequence $\sigma_t$. For each $\sigma_t$, $t=1,2, ...$, the target problem with the objective function $F_{\sigma_t}$ is solved initializing the solver with an initial point corresponding to the solution calculated at the previous step $t-1$. One would expect the algorithm to approach the actual optimizer of \ref{P0} for small values of $\sigma>0$, which yields a good approximation of the $\ell_0$ norm. More technically, the SL0 method has been proven to converge to the sparsest solution with a certain choice of the parameters, under some sparsity constraint expressed in terms of Asymmetric Restricted Isometry Constants \cite{Hosein2010}, that are in practice two distinct constants appearing, respectively, in the first and the second inequality of the RIP.

Another representative of the relaxation based techniques is the LiMapS algorithm \cite{adamo2017,adamo2011fixed}, which consists in an iterative method based on Lipschitzian mappings that, on the one hand promote sparsity and on the other hand restore the feasibility condition of the iterated solutions. Specifically, LiMapS adopts a nonlinear parametric family of shrinkage functions $f_\lambda(\a) = \a(1-e^{-\lambda|\a|})$, $\lambda>0$, acting on the iterate's coefficients in order to drive the search toward highly sparse solutions. Then it applies an orthogonal projection to map the obtained near-feasible point onto the affine space of solutions for $\Phi\a=s$. The combination of these two mappings induces the iterative system to find the solution in the subspace spanned by as small as possible number of dictionary atoms. The LiMapS algorithm has been shown to converge to minimum points of a relaxed variant of \ref{P0} defined using the sparsity promoting functions $f_\lambda$, when the parameter $\lambda>0$ is scheduled as a suitable increasing sequence having a sufficient rate of growth and some positive definiteness condition of a Hessian matrix is satisfied \cite{adamo2017}. Moreover, in the noisy model case, $s=\Phi\a + \varepsilon$, the distortion thus introduced into the generated solution is bounded as $O(\|\varepsilon\|)$ with a constant depending on the lower frame bound of $\Phi$.

%====================================================
\section{Phase Transition in Sparse Recovery}
%====================================================
Many physical processes show qualitative behaviors that are extremely different when some parameter(s) of the process trespass a certain structural threshold or boundary. Similar phenomena occur in many other natural sciences as well as in many branches of applied and pure mathematics, such as global characteristics emerging in randomly generated graphs \cite{bollobas2007phase} or the existence and complexity of solutions to constraint satisfaction problems in logic \cite{zhang2001phase}, to cite a few. Interestingly for sparse models, many problems and corresponding algorithms of sparse recovery exhibits this behavior too.

In order to quantitatively illustrate such phenomenon by comparing several well-known sparse optimization methods in literature, we adopt the experimental analysis proposed in \cite{donoho2009observed}. Specifically, Donoho and Tanner demonstrated that, assuming the solution to \ref{P0} is $k$-sparse, and the dimensions/parameters $(k,n,m)$ of the linear problem are large, the capability of many sparse recovery algorithms indeed are expressed by the phenomenon of phase transition.

%According to that protocol, we provide the phase spaces in terms of 
%Signal-to-Noise-Ratio, i.e., $\text{SNR} = 20\log_{10}\|\alpha\|/\|\alpha-\alpha^*\|$, and working 
%them out varying $\delta=\frac{n}{m}$ and $\rho=\frac{k}{n}$, that is the normalized measure of 
%problem indeterminacy\index{indeterminacy} and the normalized measure of the sparsity, respectively.

According to this analysis, using randomly generated instances of the matrix $\Phi$ and true $k$-sparse vector $\a^*$, we build instances $(\Phi,s)$ of \ref{P0} such that $\Phi\a^*=s$. We experimentally show that the methods we consider here exhibit a phase transition by measuring the Signal-to-Noise-Ratio between $\a^*$ and the recovered solution $\a$, i.e. $\text{SNR} = 20\log_{10}\|\alpha\|/\|\alpha-\alpha^*\|$, measured in dB units. In particular, the elements of atoms collected in matrix $\Phi$ are i.i.d. random variables drawn from standard Gaussian distribution, while sparse coefficients $\a^*$ are randomly generated by the so-called Bernoulli-Gaussian model. Let $\omega = (\omega_1, ...,\omega_m)$ be a vector of i.i.d. standard Gaussian variables and $\theta = (\theta_1, ..., \theta_m)$ be a vector of i.i.d. Bernoulli variables with parameter $0\leq \rho\leq 0.5$. The Bernoulli-Gaussian vector $\a^* = (\a_1^*, ..., \a_m^*)$ is then given by $\a_i^* = \theta_i\cdot\omega_i$, for all $i = 1,...,m$. Regarding the instance size, we fix $n = 100$, and we let the sparsity level $k$ and the number of unknowns $m$ range in the intervals $[1,50]$ and $[101,1000]$, respectively. The SNR is achieved by averaging over 100 randomly generated trials for every $\delta = \frac{n}{m}$ and $\rho = \frac{k}{n}$, that are the normalized measure of problem indeterminacy\index{indeterminacy} and the normalized measure of the sparsity, respectively.

%Precisely, in all tests, the frames $\Phi$ and the true coefficients $\alpha^*$ are randomly generated using the noiseless 
%Gaussian-Bernoulli stochastic model, i.e., all $\Phi_{ij} \sim \mathcal{N}(0,n^{-1})$ and 
%$\alpha_i^* \sim p_i\cdot\mathcal{N}(0,\sigma)$ are i.i.d variables, for all $i,j\in\{1,\dots,m\}$ 
%and $p_i$ is a fixed probability. We set $n=100$, and we let sparsity level $k$ and the number of 
%unknowns $m$ range in the intervals $[1,100]$ and $[101,1000]$ respectively. For each pair $(k,m)$, 
%we perform $100$ trials, then we average on the achieved SNRs.   

\begin{figure*}[h!]
	\centering
	\includegraphics[width=0.97\textwidth]{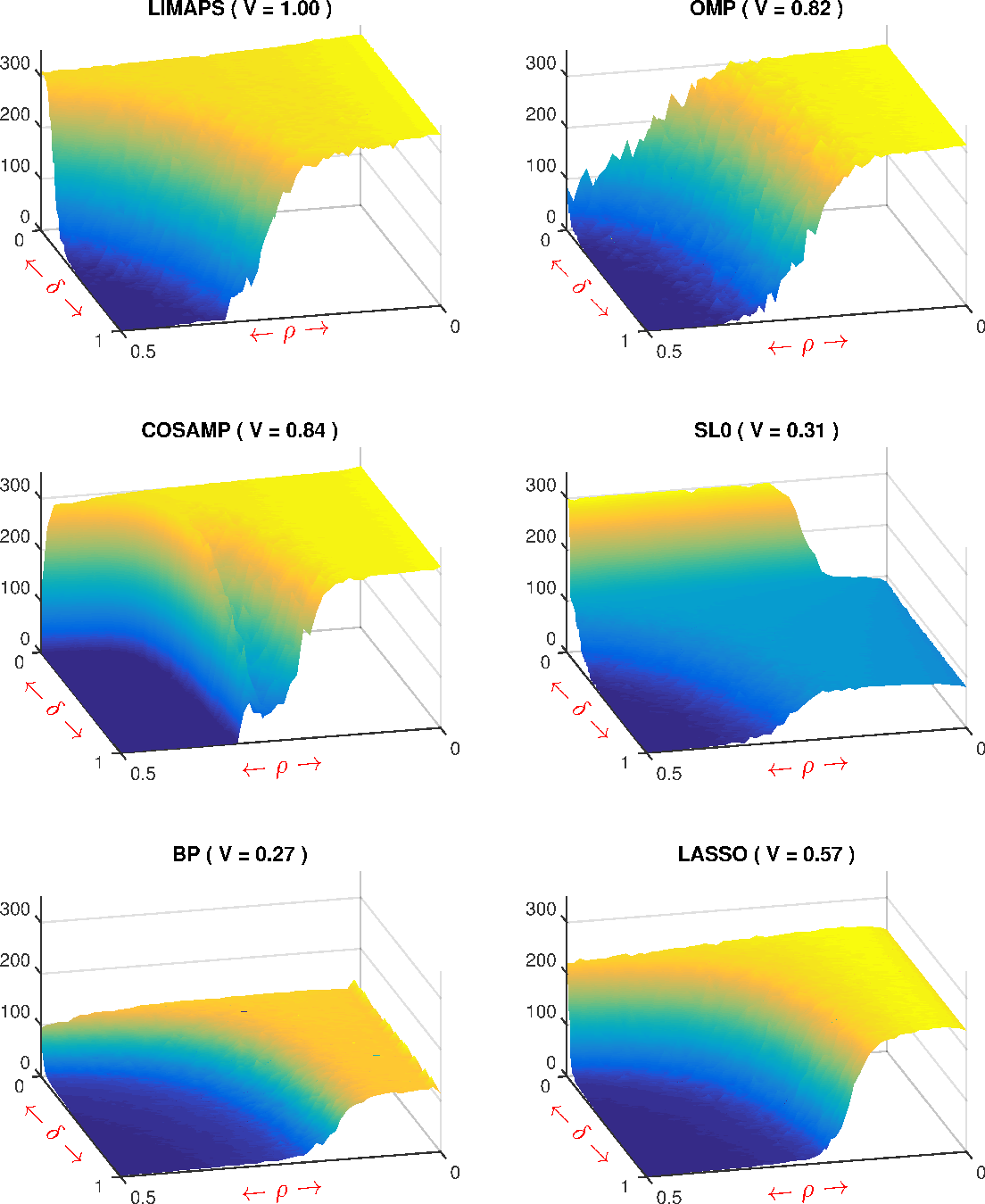}
	\caption{SNR of phase transitions of both $\ell_0$-minimizers  (first two rows) and 
		$\ell_1$-minimizers (third row) methods. The domain is defined by $(\delta,\rho)\in[0, 
		1]\times[0, 0.5]$. Next to the method name, $V$ represents the volume under the surface 
		normalized to that of LiMapS.}
	\label{phase}
\end{figure*}

In Figure \ref{phase} we report the 3D phase transitions on some well-known methods. Specifically, we refer to both  $\ell_0$-norm targeted methods such as,  OMP \cite{Tropp07OMP}, 
CoSaMP \cite{NT09}, LiMapS \cite{adamo2017} and  SL0  \cite{mohimani2008fast}, as well as to the $\ell_1$-norm targeted methods Lasso \cite{Tr02} and BP \cite{CDS98,Candes2005}.
The image clearly show the existence of a sharp phase transitions or a 
``threshold'' that partitions  the phase space into a  \textit{recoverable} region, where it is 
possible to achieve a vanishing reconstruction-error probability, from an \textit{unrecoverable} 
region in which a large error probability will eventually approach to one. The latter case corresponds 
to high sparsity measures, and low problem indeterminacy.  Qualitatively, the LiMapS algorithm reached the best results in the experiments, having the largest area of high recoverability. A quantitative assessment criterion is provided by the volume $V$ under the surface, computed by summing up the SNRs of each method in correspondence of the discrete mesh in the $\delta$-$\rho$ plane. These measures, normalized dividing by that $V$ value of the best performing algorithm, are reported in Figure~\ref{phase}, next to the method's name.
The simulations were performed using publicly available MATLAB implementation of the algorithms\footnote{SparseLab from Stanford University at \url{http://web.stanford.edu/group/sparselab}, SL0 from \url{http://ee.sharif.edu/~SLzero}, LiMapS from \url{https://phuselab.di.unimi.it/resources.php} and CoSaMP from \url{http://mathworks.com/matlabcentral}}.

\
%====================================================
\section{Sparse Dictionary Learning}
%====================================================
In the problems studied in previous sections we were interested in well representing the signal $s$ with a given dictionary $\Phi$ under a parsimony postulate. One of course awaits that the fidelity of this representation highly depends on the characteristics of the dictionary through the formal properties studied above. These in turn should also affect the level of sparsity in the representation of the data, that however can feature extreme variability. Such variability suggests that the design of suitable dictionaries that adaptively capture the features underlying the data is a key step in building machine learning models.

In literature, the proposed methods of \textit{dictionary design} can be classified into two types \cite{rubinsteinelad10}. 
The former consists in building \textit{structured dictionaries} generated from analytic prototype signals. For instance, these comprise dictionaries formed by set of time-frequency atoms such as window Fourier frames and Wavelet frames \cite{daubechies92}, adaptive dictionaries based on DCT \cite{guleryuz06a}, Gabor functions \cite{mallatzhang93}, bandelets \cite{lepennecmallat05} and shearlets \cite{easleyetal08}. 

The latter type of design methods arises from the machine learning {field} {and}  consists in \textit{training a dictionary} from available signal examples, that turns out to be more adaptive and flexible for the considered data and task. The first approach in this sense \cite{olshausenfield97} proposes a statistical model for natural image patches and searches for an overcomplete set of basis functions (dictionary atoms) maximizing the average log-likelihood (ML) of the model that best accounts for the images in terms of sparse, statistically independent components. In \cite{kreutzetal03}, instead of using the approximate ML estimate, a dictionary learning algorithm is developed for obtaining a Bayesian MAP-like estimate of the dictionary under Frobenius norm constraints. The use of Generalized Lloyd Algorithm for VQ codebook design suggested the iterative algorithm named MOD (Method of Optimal Directions) \cite{enganetal99}. It adopts the alternating scheme, first proposed in \cite{enganetal98}, consisting in iterating two steps: signal sparse decomposition and dictionary update. In particular, MOD carries out the second step by adding a matrix of vector-directions to the actual dictionary. 

Alternatively to MOD, the methods that use least-squares solutions yield optimal dictionary updating, in terms of residual error minimization. For instance, such an optimization step is carried out either iteratively in ILS-DLA \cite{enganetal07} on the whole training set (i.e., as batch), or recursively in RLS-LDA \cite{skrettingengan10} on each training vector (i.e., continuously). In the latter method the residual error includes an exponential factor parameter for forgetting old training examples. With a different approach, K-SVD \cite{eladaharon06} updates the dictionary atom-by-atom while re-encoding the sparse non-null coefficients. This is accomplished through rank-1 singular value decomposition of the residual submatrix,  accounting for all examples using the atom under consideration. Recently, Sulam et al. \cite{Sulam2016} introduced OSDL, an hybrid version of dictionary design, which builds dictionaries, fast to apply, by imposing a structure based on a multiplication of two matrices, one of which is fully-separable cropped Wavelets and the other is sparse, bringing to a double-sparsity format. Another method maintaining the alternating scheme is the R-SVD \cite{grossi2017orthogonal}, an algorithm for dictionary learning in the sparsity model, inspired by a type of statistical shape analysis, called  Procrustes method\footnote{Named after the ancient Greek myth of Damastes, known as Procrustes, the  ``stretcher,'' son of Poseidon, who used to offer hospitality to the victims of his brigandage compelling them to fit into an iron bed by stretching or cutting off their legs.} \cite{gowerdijksterhuis04}, which has applications also in other fields such as psychometrics \cite{schoenemann66} and crystallography \cite{kabsch76}. In fact, it consists in applying Euclidean transformations to a set of vectors (atoms in our case) to yield a new set with the goal of optimizing the model fitting measure.

%====================================================
\subsection{Algorithms based on alternating scheme}
%====================================================
To formally describe the dictionary learning problem we use the notation $A=\{a_i\}_{i=1}^q\in\R^{p\times q}$ to indicate a $p\times q$ real-valued matrix with columns $a_i\in\R^p, i=1,...,q$. Suppose we are given the training dataset $Y = \{y_i\}_{i=1}^L\in\R^{n\times L}$. The sparse dictionary learning problem consists in finding an overcomplete dictionary matrix  $D=\{d_i\}_{i=1}^m\in\R^{n\times m}$ ($n<m$), which minimizes the least squares errors $\|y_i-Dx_i\|_2^2$, so that all coefficient vectors $x_i\in\R^m$ are $k$-sparse.
Formally, by letting $X=\{x_i\}_{i=1}^L\in\R^{m\times L}$ denote the coefficient matrix, this problem can be precisely stated as 
\begin{equation}\label{eq:learningproblem}
\underset{D\in\R^{n\times m},X\in\R^{m\times L}}{\argmin} \|Y-DX\|_F^2 \quad \text{subject to}
\quad \|x_i\|_0\leq k,\quad i=1,...,L.
\end{equation}
One can multiply the $i$-th column of $D$ and divide the $i$-th row of $X$ by a common non-null constant to obtain another solution attaining the same value.  Hence, w.l.o.g. atoms in $D$ are constrained to be unit $\ell_2$-norm, corresponding to vectors $d_i$ on the unit $(n-1)$-sphere $\mathbb S^{n-1}$ centered at the origin.

The search for the optimal solution is a difficult task due both to the combinatorial nature of the problem and to the strong non-convexity given by the $\ell_0$ norm conditions. We can tackle this problem adopting the well-established alternating variable optimization scheme \cite[\S 9.3]{nocedalwright2006}, which consists in repeatedly executing the two steps:
\begin{description}
	\item[{Step 1.}] Sparse coding: solve problem \eqref{eq:learningproblem} for $X$ only (fixing the  
	dictionary $D$)
	\item[{Step 2.}] Dictionary update: solve problem \eqref{eq:learningproblem} for $D$ only (fixing $X$).
\end{description}
In particular, for sparse decomposition in Step 1 one can adopt the different classes of sparse recovery algorithms: BP, Lasso, LiMapS, SL0, and often OMP is applied because of its simplicity. A well designed sparse dictionary learning algorithm should be weakly affected by this choice.
Step 2 represents the core step of the learning process for a dictionary to be representative of the data $Y$. Let us view how two alternating scheme based methods perform this step.

%====================================================
\subsection{R-SVD}
%====================================================
The Procrustes analysis is the technique applied in R-SVD algorithm \cite{grossi2017orthogonal}: it consists in applying affine transformations (shifting, stretching and 
rotating) to a given geometrical object in order to best fit the shape of another target object. When the admissible transformations are restricted to orthogonal ones, it is referred to as Orthogonal Procrustes analysis \cite{gowerdijksterhuis04}. 

Basically, in R-SVD, after splitting the dictionary $D$ into atom groups, the Orthogonal Procrustes analysis is applied to each group to find 
the best rotation (either proper or improper) that minimizes the total least squares error. Consequently, each group is updated by 
the optimal affine transformation thus obtained.
Formally, let $I\subset[m]$ denote a set of indices for matrix columns or rows. Given any index set $I$ of size $s=|I|$, let $D_I\in\R^{n\times s}$ be the
submatrix (subdictionary) of $D$ formed by the \textit{columns} indexed by $I$, that is $D_I=\{d_i\}_{i\in I}$, and let
$X_I\in\R^{s\times L}$ be the 
submatrix of $X$ formed by the \textit{rows} indexed by $I$; hence $s$ is the size of atom group $D_I$. 
In this setting, we can decompose the product $DX$ into the sum
$$
DX=D_I X_I + D_{I^c} X_{I^c}
$$
of a matrix $D_I X_I$ dependent on the group $I$ and a matrix $D_{I^c} X_{I^c}$ dependent on the complement $I^c=[m]\smallsetminus I$.
Therefore, the objective function in eq.\ \eqref{eq:learningproblem} can be written as
$\|Y-DX\|_F^2 = \|Y-D_{I^c} X_{I^c} - D_I X_I\|_F^2$.

Now, after isolating the term $D_IX_I$ in 
$\|Y-DX\|_F^2$ and setting $E:=Y-D_{I^c} X_{I^c}$, one can consider the optimization problem
\begin{equation} \label{eq:learningsubproblem}
\underset{S\in\R^{n\times s}}{\argmin}\|E-SX_I\|_F^2
\qquad \text{ subject to }\qquad S\subset \mathbb S^{n-1} 
\end{equation}
that corresponds to solving a subproblem of Step 2 by restricting the update to group $D_I$ of unit $\ell_2$-norm atoms.

The method aims at yielding a new atom group $S=D'_I$, in general suboptimal for problem \eqref{eq:learningsubproblem}, by an orthogonal transformation matrix $R\in O(n,\R)$ (i.e., $R^T R=I$) applied on $D_I$, namely $D'_I=RD_I$. Remind that $O(n,\R)$ is formed by proper rotations $R\in SO(n,\R)$ and improper rotations (or rotoreflections) $R\in O(n,\R)\smallsetminus SO(n,\R)$.
Therefore, the search for such an optimal transformation can be stated as the following minimization problem
\begin{equation}\label{eq:procrustes}
\underset{R\in O(n,\R)}{\min}\ \|E-RH\|_F^2
\end{equation}
where $H:=D_I X_I\in\R^{n\times L}$. Notice that in denoting $E$ and $H$ we omit the dependence on $I$.
Problem \eqref{eq:procrustes} is known precisely as the \textit{Orthogonal Procrustes problem}
\cite{gowerdijksterhuis04} and can be interpreted as finding the rotation of a subspace matrix 
$H^T$ to closely approximate a subspace matrix $E^T$ \cite[\S 12.4.1]{golub96}.

The orthogonal Procrustes problem admits (at least) one optimal solution $\hat R$ which is \cite{golub96} the transposed orthogonal 
factor $Q^T$ of the polar decomposition $EH^T = QP$, and can be effectively computed as 
$\hat R=Q^T=VU^T$ from the orthogonal matrices $U$ and $V$ of the singular value decomposition $EH^T = U\Delta 
V^T\in\R^{n\times n}$.
Hence the rotation matrix sought is $\hat R = VU^T$, the new dictionary $D'$ has the old columns 
of $D$ in the positions $I^c$ and the new submatrix $D'_I=\hat RD_I$ in the positions $I$, while 
the new non-increased value of reconstruction error is
$$
\|Y-D'X\|_F^2=\|Y-D_{I^c}X_{I^c}-VU^T D_I X_I\|_F^2
\leq \|Y-DX\|_F^2.
$$
At this point the idea of the whole R-SVD algorithm is quite straight-forward:
\begin{enumerate}
	\item at each dictionary update iteration (Step 2) partition the set of column indices $[m]=I_1\sqcup I_2\sqcup \cdots \sqcup I_G$ into $G$ subsets,
	\item then split $D$ accordingly into atom groups $D_{I_g}$, $g=1,...,G$, and
	\item update every atom group $D_{I_g}$.
\end{enumerate}

These updates can be carried out either in parallel or sequentially with some order: for example, the sequential update with ascending order of atom popularity, i.e. sorting the indices $i\in [m]$ w.r.t. the usage of atom $d_i$, computable as $\ell_0$-norm of the $i$-th row in $X$.
For sake of simplicity one can set the group size uniformly to $s = |I_g|$ for all $g$, possibly except the last group ($G = \lceil m/s \rceil$) if $m$ is not a multiple of $s$: $|I_G|=m-Gs$.
Other grouping criteria could be adopted: eg. random balanced grouping, Babel function \cite{tropp2004greed} (also called cumulative coherence, a variant alternative to mutual coherence) based partitioning, and clustering by absolute cosine similarity.

After processing all $G$ groups, the method moves to the next iteration, 
and goes on until a stop condition is reached, e.g. the maximum number of iterations as commonly
chosen, or an empirical convergence criterion based on distance between successive iterates.
The main steps are outlined\footnote{The Matlab code implementing the algorithm is available on the website \url{https://phuselab.di.unimi.it/resources.php}} in Algorithm \ref{alg:K-SVD}.
\begin{algorithm}
	\caption{R-SVD}
	\label{alg:K-SVD}
	\begin{algorithmic}[1]
		\REQUIRE $Y\in\R^{n\times L}$: column-vector signals for training the dictionary
		\ENSURE $D\in\R^{n\times m}$: trained dictionary; $X\in\R^{m\times L}$: sparse encoding of $Y$
		\STATE Initialize dictionary $D$ picking $m$ examples from $Y$ at random
		\REPEAT
		\STATE Sparse coding: $X=\argmin_{X} \|Y-DX\|_F^2$ subject to $\|x_i\|_0\leq k$ for $i=1,...,L$
		\STATE Partition indices $[m]=I_1\sqcup I_2\sqcup ... \sqcup I_G$ sorting by atom popularity
		\FOR{$g=1,...,G$}
		\STATE $J=I_g$
		\STATE $E = Y-D_{J^c} X_{J^c}$
		\STATE $H=D_J X_J$
		\STATE $R=\argmin_{R\in O(n)} \| E-R H\|_F^2=VU^T$ by rank-$s$ SVD $EH^T=U\Sigma V^T$
		\STATE $D_J=RD_J$
		\ENDFOR
		\RETURN $D,X$
		\UNTIL{stop condition}
\end{algorithmic}
\end{algorithm}
Notice that in R-SVD the renormalization of atoms to unit length at each	iteration is not necessary since they are inherently yielded with such a condition from the Procrustes analysis, and hence in practice some renormalizing computations as in ILS-DLA \cite{enganetal07} and K-SVD \cite{aharonetal06} can be avoided.

%====================================================
\subsection{K-SVD}
%====================================================
The K-SVD algorithm still performs an alternating optimization scheme, but the dictionary update step is carried out through many rank-1 singular value decompositions, which justify the name. Precisely, recall the decomposition of $DX$ into the sum $DX=D_I X_I + D_{I^c} X_{I^c}$ introduced for R-SVD. If we choose the singleton atom $I=\{h\}$, i.e. $d_h$, we can consider the index set $\omega(h)=\{\ell\in [L]: X_{h,\ell}\neq 0\}$ indicating the examples $y_\ell, \ell\in\omega(h)$, that use the atom $d_h$ in the approximate representation of $Y$ by $DX$.
The error matrix in this approximate representation must be $Y_{\omega(h)}-D\tilde X$, where $\tilde X$ is the submatrix of $X$ formed by the columns (indexed by) $\omega(h)$. Taking $\tilde Y:=Y_{\omega(h)}$, i.e. the columns $\omega(h)$ of $Y$, we have:
$$
\tilde Y - D\tilde X = \tilde Y - D_{[m]\smallsetminus\{h\}} \tilde X_{[m]\smallsetminus\{h\}} -  d_h \tilde x_h = E_h - d_h \tilde x_h
$$
with the obvious definition of $E_h$, where $\tilde x_h$ is the $h$-th row of $\tilde X$. The last term $d_h \tilde x_h\in\R^{n\times L}$ is a rank-1 matrix. The K-SVD then updates the atom $d_h$ and the encoding row vector $\tilde x_h$ by minimizing the squared error:
$$
\min_{d\in\R^{n\times 1}, x\in\R^{1\times \#\omega(h)}} \|E_h - d x\|_F^2
$$
which is indeed a rank-1 approximation problem, that can be easily solved by a truncated SVD of $E_h = U \Delta V^T$. The new atom $d_h'$ results to be the first column of $U$, while the relative encoding coefficients, $\tilde x_h'$, are the first column of $V$. It is easy to see that the columns of $D$ remain normalized and the support of all representations either stays the same or gets smaller \cite{aharonetal06}.

The above update process is repeated for every choice $h=1,...,m$ of an atom $d_h$ in the dictionary update step, and the two alternating steps are iterated until a certain convergence criterion is satisfied in the whole K-SVD algorithm.

%====================================================
\subsection{Dictionary learning on synthetic data}
%====================================================
For demonstrating a practical application, we apply the sparse dictionary learning method on synthetic data conducting empirical experiments with both R-SVD and K-SVD using OMP as sparse recovery algorithm.
Following \cite{aharonetal06}, the true dictionary $D \in \R^{n\times m}$ is randomly drawn, 
with i.i.d. standard Gaussian distributed entries and each column normalized to unit $\ell_2$-norm. 
The training set $Y\in \R^{n\times L}$ is generated column-wise by $L$ linear combinations of $k$ 
dictionary atoms selected at random, and by adding i.i.d. Gaussian entry noise matrix $N$ with various noise power expressed as SNR, i.e. $Y = DX + N$. We measure the performances of the K-SVD and R-SVD algorithms in terms of the reconstruction error (or quality) expressed as 
$\ESNR=20 \log_{10} (\|Y\|_F / \|Y-\tilde D \tilde X\|_F)$ dB, where $\tilde D$ and $\tilde X$ are the learned dictionary and the sparse encoding matrix, respectively.

We consider dictionaries of size $50\times 100$ and $100\times 200$, 
dataset of size up to $L = 10000$ and 
sparsity $k = \{5,10\}$. The algorithms K-SVD and R-SVD are run for $T=200$ dictionary update 
iterations, that turns out to be sufficient to achieve empirical convergence of the 
performance measure. For each experimental setting we report the average error over 100 trials.

% FIGURE 2
\begin{figure}[h]
	%	\begin{adjustwidth}{-1.5in}{0in}
	\centering
	\includegraphics[width=0.49\linewidth]{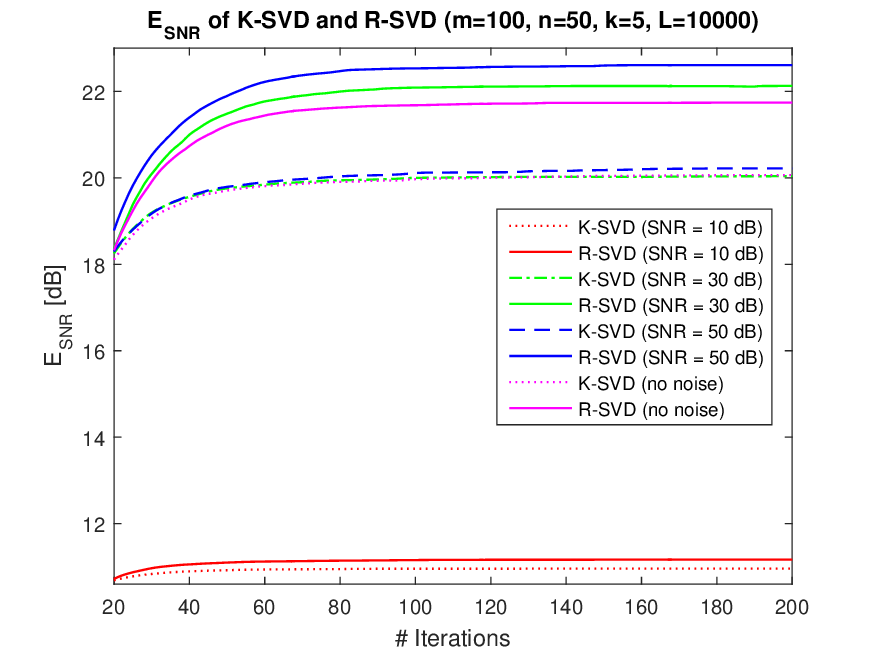}
	\includegraphics[width=0.49\linewidth]{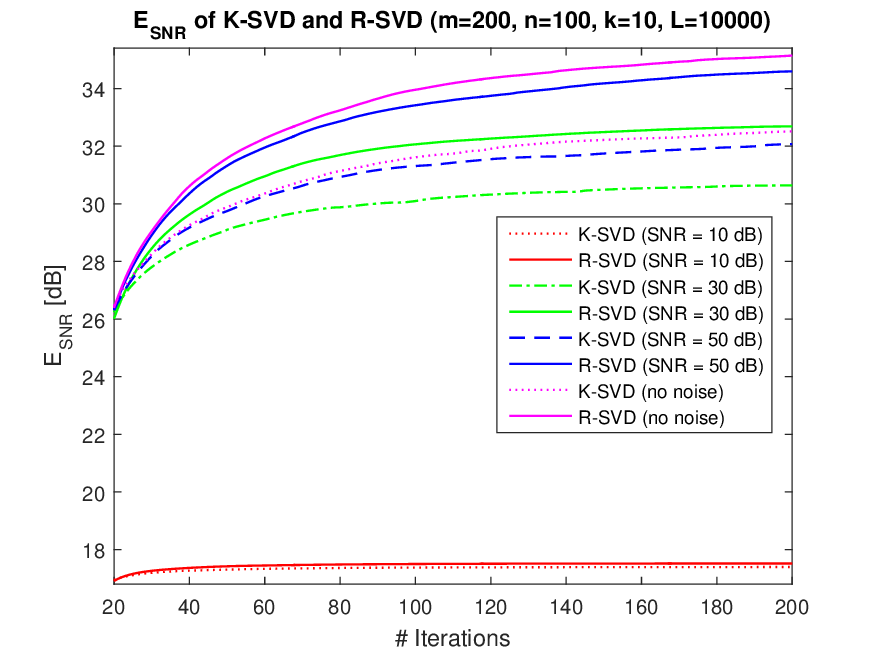}
	\caption{Average reconstruction error $\ESNR$ in sparse representation using dictionary learnt by K-SVD (non-solid lines) and R-SVD (solid lines), for $L=10000$ synthetic vectors varying the additive noise power (in the legend). Averages are calculated over 100 trials and plotted versus update iteration count.  \textit{Left}:  $D \in \R^{50 \times 100}$ with sparsity $k=5$,   \textit{Right}:   $D \in \R^{100 \times 200}$ with sparsity $k=10$.}
	\label{fig:SyntResults}
	%	\end{adjustwidth}
\end{figure}
In Fig.\ \ref{fig:SyntResults}~we highlight the learning trends of the two methods, plotting at each 
iteration count the $\ESNR$ values on synthetic vectors $Y=DX+N$, varying the additive noise power level, SNR $= 
10, 30, 50, \infty$ (no 
noise) dB. It can be seen that, after an initial transient,  the gap between R-SVD and K-SVD 
increases with the iteration count, establishing a final gap of 2 dB or more in conditions of middle-low noise power (SNR $\geq 30$ dB).

% FIGURE 3
\begin{figure}[h!]
	%	\begin{adjustwidth}{-2.25in}{0in}
	\centering
	\includegraphics[width=0.8\linewidth]{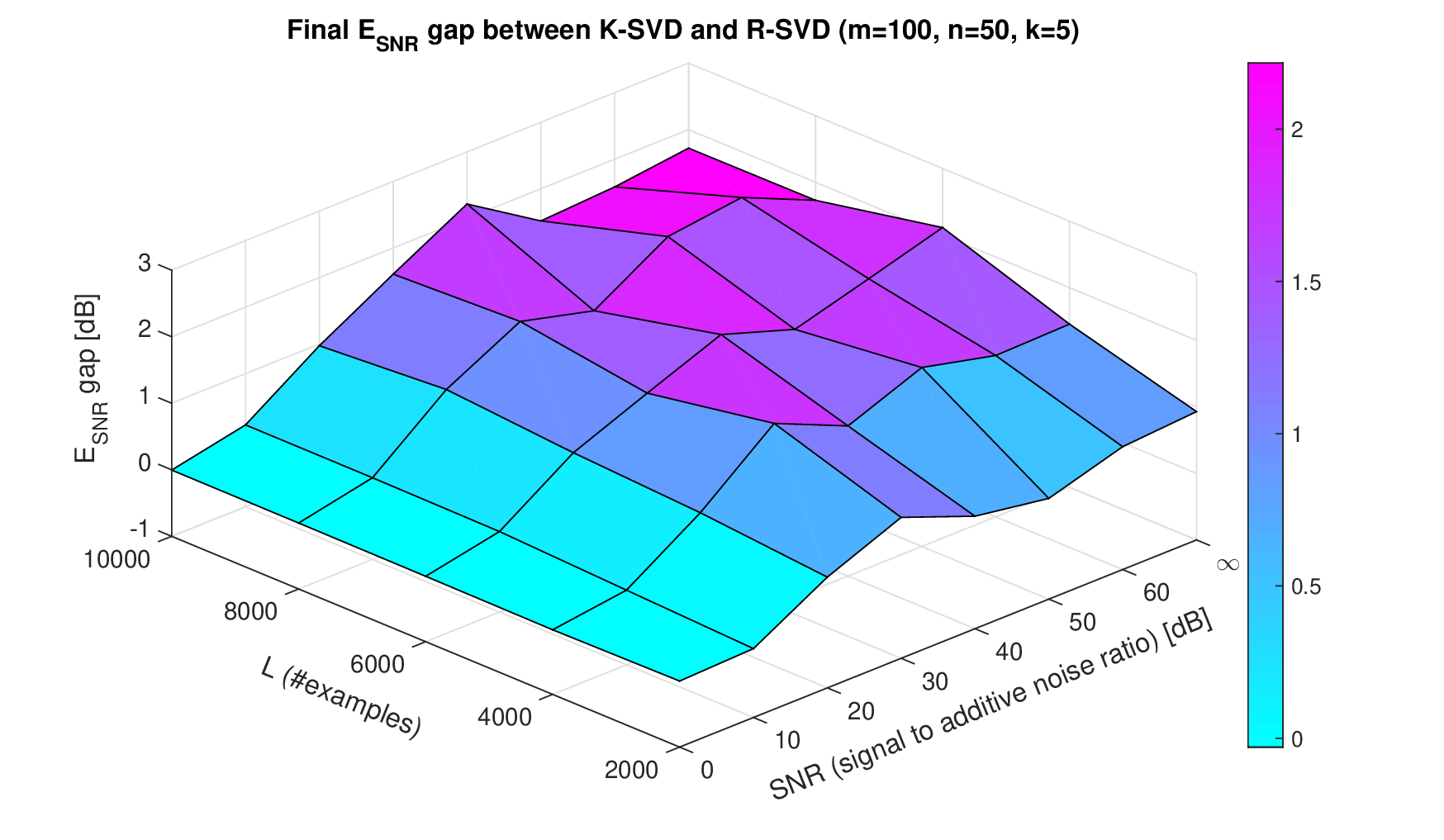}
	\includegraphics[width=0.8\linewidth]{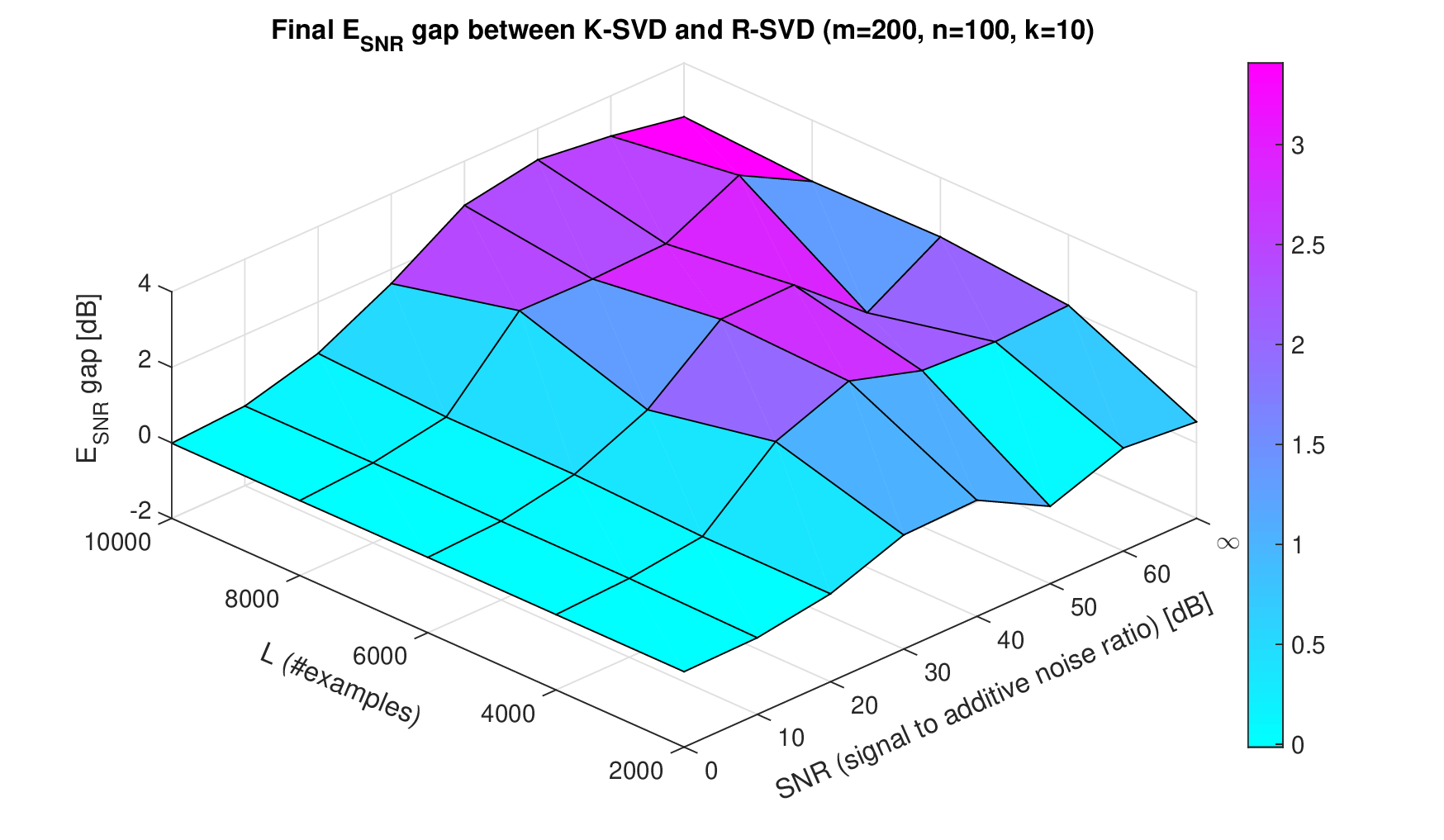}
	\caption{Gap between final ($T=200$) $\ESNR$ of K-SVD and R-SVD obtained with all parameter combinations $L=2000$, 4000, 6000, 8000, 10,000 and SNR $= 0,10,20,30,40,50,60, \infty$ (no noise). Results are averages over 100 trials; points are interpolated with colored piece-wise planar surface for sake of readability. \textit{Top}: $D\in\R^{50\times 100}$ with sparsity $k=5$. \textit{Bottom}: $D\in\R^{100\times 200}$ with sparsity $k=10$.}
	\label{fig:SyntSNRgap}
	%	\end{adjustwidth}
\end{figure}

In order to explore the behavior of R-SVD and K-SVD in a fairly wide range of parameter values, we 
report in Fig.\ \ref{fig:SyntSNRgap}~the gaps between their final ($T=200$) reconstruction error 
$\ESNR$, varying $L$ in $2000\div 10000$, noise power level SNR in $0\div 60$ dB, and in case of no noise. 
Dictionary sizes, sparsity and number of trials are set as above.
\begin{table}[b]
	\centering
	\renewcommand{\arraystretch}{1} %reduce space between rows
	\caption{Average number of atoms correctly recovered (matched) by K-SVD and R-SVD algorithms at various SNR levels of additive noise on dictionary $D$ of size  $50\times 100$ and  $100\times 200$. $L=10000$, and remaining parameter values as in Fig. \ref{fig:SyntSNRgap}.}
	\label{tab:recoveredatoms}
	\resizebox{1\textwidth}{!}{
		\begin{tabular}{@{}ccccccccc@{}}
			\toprule
			\multicolumn{9}{c}{\textbf{Number of recovered atoms}}\\
			\toprule
			& \multicolumn{2}{c}{{SNR = 10 dB}}   & \multicolumn{2}{c}{{SNR = 30 dB}}  & \multicolumn{2}{c}{{SNR = 50 dB}} & \multicolumn{2}{c}{{No noise}} \\
			
			\cmidrule(lr){2-3}
			\cmidrule(lr){4-5}
			\cmidrule(lr){6-7}
			\cmidrule(lr){8-9}
			${n\times m}$   & {K-SVD}      & {R-SVD}     & {K-SVD}      & {R-SVD}     & {K-SVD}      & {R-SVD}     & {K-SVD}    & {R-SVD}    \\
			$50\times 100$     & 94.52    & 97.37   & 92.15    & 94.08   & 92.1      & 93.84    & 92.07    & 94.03   \\
			$100\times 200$   & 195.82  & 199.02     & 192.42         & 194.98      & 192.49         & 194.57  & 192.87        & 194.7  \\ \bottomrule
		\end{tabular}
	}
\end{table}
When the additive noise power is very high (e.g., SNR $=0$ or 10 dB) the two methods are practically 
comparable: the presence of significant noise could mislead most learning algorithms. On the other hand, when the noise is quite low the R-SVD algorithm performs better than K-SVD.
Another interesting empirical investigation is the evaluation of the number of correctly identified atoms in order to measure the ability of the learning algorithms in recovering the original dictionary $D$ from the 
noise-affected data $Y$. 
This  is accomplished by maximizing the matching between true atoms $d_i$ of the original dictionary 
and atoms $\tilde d_j$ of the dictionary $\tilde D$ yielded by an algorithm: two unit-length atoms 
$(d_i,\tilde d_j)$ are considered matched when their cosine dissimilarity is small \cite{aharonetal06}, 
i.e. precisely
$$
1 - |d_i^T \tilde d_j| < \varepsilon:=0.01.
$$
In Table \ref{tab:recoveredatoms} we report the average number of atoms recovered by the K-SVD and R-SVD algorithms on randomly initialized instances at various additive noise power levels.

\bibliographystyle{abbrv}
\bibliography{biblio,Adamo,TCS,TETC,PLOS-ONE}

\end{document}